\RequirePackage{fix-cm}
\documentclass[]{article}
\usepackage{arxiv}
\usepackage[english]{babel}
\usepackage{lipsum}
\usepackage{graphicx} 
\usepackage{placeins}
\usepackage{bm}
\usepackage{booktabs} 
\usepackage{csquotes}
\usepackage{subfigure}
\usepackage{multirow}
\usepackage{listings}
\usepackage{xcolor}
\definecolor{abstract_background}{RGB}{235,235,235}
\usepackage{float}
\usepackage{color, colortbl}
\definecolor{ao(english)}{rgb}{0.0, 0.5, 0.0}
\usepackage{xr-hyper}
\usepackage{hyperref}
\hypersetup{colorlinks,linkcolor={blue},citecolor={ao(english)},urlcolor={red}} 
\usepackage{stackrel,amssymb}
\usepackage{amsfonts,amsmath,amstext,amsbsy,amsthm}
\theoremstyle{definition}

\newtheorem{definition}{Definition}[section]
\newtheorem{theorem}{Theorem}[section]

\newtheorem{lemma}[theorem]{Lemma}

\usepackage{cleveref}
\usepackage{amsmath}
\definecolor{LightCyan}{rgb}{0.88,1,1}
\definecolor{LightRed}{rgb}{1,0.7,0.7}

\newcommand{\R}{\mathbb{R}}
\newcommand{\C}{\mathbb{C}}

\title{Random Projection Neural Networks of Best Approximation: Convergence theory and practical applications}

\author{
\textbf{Gianluca Fabiani}\textcolor{teal}{$^{1,}$}\thanks{email: \texttt{gianluca.fabiani@unina.it}}\hspace{0.1cm}
{}\\
\textcolor{teal}{$^{(1)}$} Modelling Engineering Risk and Complexity, \emph{Scuola Superiore Meridionale}, Naples 80138, Italy \hspace{1cm}\\
}

\usepackage{amsopn}

\begin{document}

\maketitle
\begin{abstract}
\colorbox{abstract_background}{\begin{minipage}{1\linewidth}
We investigate the concept of Best Approximation for Feedforward Neural Networks (FNN) and explore their convergence properties through the lens of Random Projection (RPNNs). RPNNs have predetermined and fixed, once and for all, internal weights and biases, offering computational efficiency.
We demonstrate that there exists a choice of external weights, for any family of such RPNNs, with non-polynomial infinitely differentiable activation functions, that exhibit an exponential convergence rate when approximating any infinitely differentiable function.
For illustration purposes, we test the proposed RPNN-based function approximation, with parsimoniously chosen basis functions, across five benchmark function approximation problems. Results show that RPNNs achieve comparable performance to established methods such as Legendre Polynomials, highlighting their potential for efficient and accurate function approximation.
\end{minipage}
}
\end{abstract}



\section{Introduction}
In the pursuit of enhancing the efficiency and accuracy of complex numerical and optimization problems, artificial neural networks (ANNs) have emerged as a powerful tool demonstrating significant improvements in real-world applications, including (but not limited to), medical science, finance, management and security \cite{abiodun2018state,ferdiana2020systematic,lim2022unfolding,jha2020doubleu,karniadakis2021physics}.
Widely used in Machine Learning (ML) tasks, including image classification, signal analysis, speech recognition, text-to-image synthesis and language processing \cite{yadav2015introduction,nassif2019speech,wolf2020transformers,gui2021review,ding2021cogview,han2022survey}, the celebrated universal approximation theorem \cite{cybenko1989approximation, hornik1990universal,leshno1993multilayer,barron1993universal,irie1988capabilities,funahashi1989approximate,carroll1989construction} has significantly influenced and propelled the adoption of ANN-based function approximation for solving various numerical analysis and mathematical modelling problems, including the solution of ordinary and partial differential equations (ODEs and PDEs) \cite{lee1990neural,meade1994numerical,gerstberger1997feedforward,lagaris1998artificial,han2018solving,lu2021deepxde,raissi2019physics,fabiani2021numerical,fabiani2023parsimonious,calabro2021extreme,dong2021computing,dong2021local}, the data-driven identification of dynamical systems \cite{rico1992discrete,krischer1993model,gonzalez1998identification,alexandridis2002modelling,raissi2018deep,raissi2019physics,karniadakis2021physics,galaris2022numerical,fabiani2023tasks,lee2023learning,dietrich2023learning}, and the physics-informed solution of functional equations \cite{yecsildirek1995feedback,levin1996control,karniadakis2021physics,lu2021learning,kalia2021learning,alvarez2023discrete,patsatzis2023slow}.\par
Since the 1990s, there has been a notable increase in research focusing on solving numerical analysis problems. Lee and Kang (1990)~\cite{lee1990neural} employed a Hopfield Neural Network to address the numerical solution of differential equations. Rico-Martinez et al. (1992) \cite{rico1992discrete} introduced a recursive multilayer ANNs architecture, emulating the implementation of the 4th-order Runge-Kutta scheme, for the identification of continuous-time ODEs and the construction of their bifurcation diagram from experimental data. Yeşildirek and Lewis (1995) \cite{yecsildirek1995feedback} employed a neural network-based controller for the feedback linearization of dynamical systems.  
Gerstberger and Rentrop (1997) \cite{gerstberger1997feedforward} utilized feedforward neural networks (FNNs) addressing stiff ODEs, as well as Differential-Algebraic Equations (DAEs).
Lagaris et al. (1998) \cite{lagaris1998artificial} systematically investigate the use of FNNs for the solution of linear and nonlinear differential equations, addressing a range of scenarios from initial and boundary value problems. 
In recent times, there has been a surge in research activity in the field, thanks to advancements in both theoretical and applied mathematics, computer science, and increased computational power. These progressions have paved the way for further developments, particularly in the application of Physics-Informed (Deep) Neural Networks (PINNs) \cite{raissi2019physics,fabiani2021numerical,karniadakis2021physics,lu2021deepxde,fabiani2023parsimonious,alvarez2023discrete}.\par

Given the myriad of numerical applications where ANNs demonstrate their efficacy, it is essential to delve into the theoretical underpinnings that substantiate their widespread adoption. 
As we transition from practical applications to theoretical insights, it becomes apparent that the success of ANNs in diverse domains is not merely coincidental but is deeply rooted in rigorous mathematical principles. Indeed, contrary to the perception of neural networks as ``magical" black boxes, there exists a substantial body of --often overlooked, underestimated, or forgotten-- theoretical advancements that underpin their effectiveness.
The universal approximation theorem \cite{gallant1988there,irie1988capabilities,cybenko1989approximation,hornik1989multilayer,carroll1989construction,funahashi1989approximate,leshno1993multilayer,barron1993universal} demonstrated that a single (and multi) hidden layer ANN possesses the ability to universally approximate, as long as the number of nodes in the hidden layer is allowed to increase without limit, any continuous function within a confined domain with arbitrary uniform accuracy. This correspond to the density property of the ANN-based functions in the space $C(\R^d)$, in the topology of uniform convergence on compacta.
Originally motivated by Kolmogorov's superposition theorem (1957) \cite{kolmogorov1957representation}, the late 1980s witnessed the arising of a multitude of results. Gallant (1988) \cite{gallant1988there} proved that ANN with the \emph{cosine squasher} activation function possess the density property and can reconstruct any Fourier series. Irie and Miyake (1988) \cite{irie1988capabilities} proved integral representation of $L^1$ functions with arbitrary plane wave kernels. 
Independently in the same year (1990), Cybenko \cite{cybenko1989approximation}, Funahashi \cite{funahashi1989approximate}, and Hornik, Stinchcombe, and White \cite{hornik1989multilayer} proved uniform convergence on compact sets with continuous sigmoidal functions, each with slightly different assumptions. Subsequently, Ito (1992) \cite{ito1992approximation} delved into the density problem using monotone sigmoidal functions with only weights of norm 1.
The conclusive result on density, as elucidated by Leshno et al. (1993) \cite{leshno1993multilayer}, asserted that the necessary and sufficient condition is the non-polynomial nature of the activation functions.
It has been established that ANNs possess the capability to simultaneously and uniformly approximate a function and multiple of its partial derivatives \cite{hornik1990universal,cardaliaguet1992approximation,ito1993approximations,attali1997approximations}, along with the corresponding density property in $C^m(\mathbb{R})$. These findings have paved the way for the application of Physics Informed (Deep) Neural Networks (PINNs) in solving differential problems and nonlinear functional equations \cite{raissi2019physics,fabiani2021numerical,karniadakis2021physics,lu2021deepxde,fabiani2023parsimonious,alvarez2023discrete}.

Recently, a multitude of findings has emerged and extended the approximation capabilities of ANNs across various function spaces. Examples include Sobolev spaces \cite{guhring2020error}, bandlimited functions \cite{montanelli2021deep}, analytic functions \cite{wang2018exponential,opschoor2022exponential}, Barron functions \cite{weinan2019barron}, and H\"older spaces \cite{shen2020deep}.
Moreover, even without a strong and comprehensive theoretical justification explicitly favoring deep neural networks (DNNs) over their shallow counterparts, empirical evidence and practical results indicate that deep learning models, with their increased complexity and capacity to capture intricate patterns, are achieving better performance in various tasks \cite{poggio2017and,schmidhuber2015deep}.\par
However, despite the remarkable empirical and theoretical achievements, there is a prevailing concern that approaches relying on DNNs currently fall short of meeting the conventional rigorous standards in terms of stability, convergence and efficiency typically required for algorithms in computational science and numerical analysis. The theoretical underpinnings of DNNs suggest their formidable capabilities to represent a wide range of functions. However, the practical application of DNNs faces challenges due to the non-convex nature of the optimization problem involved in training these networks. Theoretical guarantees often rely on idealized assumptions, and the intricate landscape of the optimization objective makes it difficult to achieve optimal results in practice.
The non-convexity of the optimization landscape introduces challenges such as saddle points, local minima, and plateaus, which can impede the convergence of optimization algorithms. Consequently, the gap between theoretical expectations and practical outcomes in DNNs remains substantial \cite{almira2021negative,adcock2021gap}.\par

%
One strategy, that aims at accelerating computations, thus proved efficient in solving challenging well-established benchmark problems of numerical analysis \cite{fabiani2021numerical,dong2021computing,dong2021local,fabiani2023parsimonious}, involves the use of Random Projection Neural Networks (RPNNs), which are essentially ANNs with predetermined and fixed internal weights and biases.
RPNNs draw inspirations from a Rosenblatt's pioneering idea, the so-called \emph{Gamba Perceptron} \cite{rosenblatt1962perceptions}, and are further motivated by Johnson-Lindenstrauss Theorem \cite{johnson1984extensions}, as well as capability to approximate any kernel distance by random features \cite{rahimi2008weighted}, and extensions of universal approximation to random fixed basis functions \cite{barron1993universal,igelnik1995stochastic,huang2006extreme}. Extensive demonstrations of this approach have shown that comparable or superior accuracy can be achieved. By circumventing the time-intensive iterative gradient-based algorithms, it also leads to a significant reduction in training time, often by thousands of times \cite{rahimi2008weighted,dong2022local,fabiani2023parsimonious,fabiani2023tasks,fabiani2021numerical,huang2015trends,dong2021local,calabro2021extreme}.
Recent studies \cite{fabiani2021numerical,calabro2021extreme} demonstrated that employing Physics-informed RPNNs, and in separate investigations incorporating local RPNNs along with domain decomposition \cite{dong2022local,dong2021local}, can outperform the classical Finite Difference (FD) and Finite Elements methods (FEM) in both accuracy and computational cost for the numerical solution of PDEs.
In Fabiani et al. (2023) \cite{fabiani2023parsimonious} it has been recently shown that the integration of RPNNs with cutting-edge numerical analysis and continuation techniques has proven superior to traditional stiff time integrators for ODEs and also Differantial Algebraic Equations (DAEs). \par

Navigating the intricate landscape of high-dimensional input spaces poses a substantial decision when considering the application of RPNNs and/or fully trainable ANNs to a given problem. As proven by Barron (1993) \cite{barron1993universal}, ANNs excel in high dimensions due to their adaptive nature, requiring fewer parameters and avoiding the \emph{curse of dimensionality}, yet they suffer from slower training. In contrast, RPNNs offer faster training but, as the input space grows, they become susceptible to the \emph{curse of dimensionality} \cite{barron1993universal}.
Nevertheless, efforts have been made to strategically and parsimoniously choose the basis functions for RPNNs, even in high-dimensional scenarios \cite{galaris2022numerical, bolager2023sampling, dong2022computing, rahimi2008weighted}.\par

In this context, our primary focus lies in assessing --both theoretically and through practical numerical tests-- the convergence rate of RPNNs, when applied to the approximation of low-dimensional infinitely differentiable functions. For this problem, a significant challenging question persists concerning the convergence rate of RPNNs as the number of neurons increases.
To the best of our knowledge, the current known convergence rates of $O(1/\sqrt{N})$ of RPNN in expectation, as given by Igelnik and Pao (1995) \cite{igelnik1995stochastic}), is not comparable with classical numerical analysis flagship methods for function approximation such as Legendre and Chebyshev polynomials that exhibit exponential convergence rate \cite{gaier1987lectures}. 
However, empirical evidence from various studies, including \cite{fabiani2021numerical,calabro2021extreme,dong2021local,dong2022local,fabiani2023parsimonious}, suggests that, especially for relatively low-dimensional problems, the effective convergence rate of RPNNs seems to surpass the expected upper bound. 

Recently, Adcock and Dexter (2021) \cite{adcock2021gap} established the existence --\emph{only theoretically}-- of a Deep Neural Network architecture and training procedure that attains the same exponential rate of convergence as compressed sensing polynomials \cite{adcock2017compressed} for analytic function approximation with the same sample complexity. However, in their conclusion they assert that ``remains a large gap between expressivity and practical performance achieved with standard methods of training".

We are of the opinion that through the exploration of RPNNs, which essentially represent ANNs trained in a distinct manner, our research can provide insights into efficient and accurate training methodologies. This, in turn, offers promising insights into the theoretical discoveries of Adcock and Dexter (2021) \cite{adcock2021gap}, giving hope for practical implementations and contributing to understanding the broader potential of ANNs for function approximation.

\subsection{Main contributions}
This work extends the ANNs convergence rate inquiry to RPNNs, presenting a deeper exploration of their connection with best approximation theory.
Here, we show that there exists always a choice of the external weights of a Feedforward neural network (FNN), with random \emph{a priori} fixed internal weights and biases and infinitely differentiable non-polynomial activation functions, that exponentially converge when approximating infinitely differentiable functions. While for the approximation of functions with less smoothness, the maximum order of differentiability poses a limit to such convergence.
To this extent we explore and extend the concept of best $L^p$ approximation \cite{davis1975interpolation,kainen1999approximation,girosi1990networks}, well-established for polynomials, to the field of ANNs.

Additionally, we introduce a novel function-informed \emph{a priori} selection of the internal weights and biases. This approach is compared with two alternatives: (a) a naive random generation \cite{huang2006extreme} and (b) a parsimonious function-agnostic choice as recommended in \cite{calabro2021extreme,fabiani2021numerical,fabiani2023parsimonious}.

We conduct numerical tests on a set of five benchmark problems for function approximation, comparing RPNNs of best $L_2$ approximation with Legendre polynomials and cubic splines. Our findings demonstrate that employing an RPNN training approach enables achieving accuracy 
comparable with the accuracy of  Legendre polynomial approximation. This surpasses the limitations encountered in classical fully trainable Deep Neural Networks (DNN) \cite{adcock2021gap}. 

Overall, our work contributes to a deeper understanding of RPNNs and their potential for efficient and accurate function approximation.

\subsection{Outline}
The paper is organized as follows. In Section \ref{sec:RPNN} we introduce the fundamentals and notations regarding RPNNs. In Section \ref{sec:best}, we prove the existence and uniqueness of the RPNN of best $L^p$ approximation. In Section \ref{sec:main} we present the main result concerning the exponential convergence rate of RPNNs of the best $L^p$ approximation. In Section \ref{sec:selection}, we give a brief overview of the possible choice for fixing internal parameters of RPNNs.
In Section \ref{sec:numerical} we perform numerical tests.
In the text, vectors will be denoted by bold symbols, and matrices will be represented by capital letters.

\section{Preliminaries on Random Projection Neural Networks}
\label{sec:RPNN}
Random Projection Neural Networks (RPNN) are a class of Artificial Neural Networks (ANNs), including Random Weights Neural Networks (RWNN) \cite{schmidt1992feed}, Random Vector Functional Link Network (RVFLN) \cite{pao1992functional,igelnik1995stochastic}, Reservoir Computing (RC)\cite{jaeger2001echo,jaeger2002adaptive}, Extreme Learning Machines (ELM) \cite{huang2006extreme,huang2015trends} and Random Fourier Features Networks (RFFN) \cite{rahimi2008weighted}. Some seed of this idea can be also found in \emph{gamba perceptron} proposed initially by Frank Rosenblatt \cite{rosenblatt1962perceptions}.\par

A fundamental work on random projections that links the above conceptually equivalent approaches is the celebrated Johnson and Lindenstrauss Lemma \cite{johnson1984extensions}, which states that there exist an approximate isometry map $\bm{F}: \R^d \rightarrow \R^k$ of input data $\bm{x}\in\R^n$ induced by a random matrix $R$:
\begin{equation}\label{JL}
\bm{F} (\bm{x}) = \dfrac{1}{\sqrt{k}} R \bm{x},
\end{equation}
where the matrix $R = [R_{ij}] \in \R^{k \times d}$ has components which are i.i.d.~random variables sampled from a normal distribution.

Notwithstanding, constructing a feature space ($x\rightarrow F(x)$) that aims at preserving Euclidean distance may not consistently constitute the optimal approach when undertaking machine learning tasks.
In this context, the effectiveness of nonlinear explicit random lifting operator, as realized by the hidden layer(s) of RPNNs, in preserving kernel distances has been investigated in detail in \cite{rahimi2008weighted}. Besides, as it has been shown (see e.g., \cite{barron1993universal,igelnik1995stochastic,gorban2016approximation}), appropriately constructed nonlinear random projections may outperform such simple linear random projections.

Let us consider a single output, single hidden layer feed-forward neural network (FNN), denoted by a function $f_N:\R^d \rightarrow\R$ with an \emph{a priori} fixed matrix of \emph{internal weights} $A \in \R^{N\times d}$ with $N$ rows $\bm{\alpha}_j \in \R^{1\times d}$ and \emph{biases} $\bm{\beta}=(\beta_1,\dots,\beta_N) \in \R^{N}$:
\begin{equation}
f_N(\bm{x};\bm{w},\beta^o,A,\bm{\beta})=\sum_{j=1}^N w_j \psi (\bm{\alpha}_j\cdot \bm{x}+ \beta_j) +\beta^{o}=\sum_{j=1}^N w_j \psi_j (\bm{x})+\beta^{o}
\label{eq:RPNN}
\end{equation}
where $N$ is the number of neurons (nodes), $d$ is the dimension of the input $\bm{x} \in \R^{d\times 1}$, $\beta^{o}$ is a constant offset, or the so-called \emph{output bias}, $\psi:\R\rightarrow\R$ is the so-called \emph{activation} (transfer) function, that for fixed parameters $\alpha_j$ and $\beta_j$ we denote as a fixed \emph{basis function} $\psi_j$, and $\bm{w}=(w_1,\dots,w_N)^{T} \in\R^{N \times 1}$ are the \emph{ external (readout) weights} that connects hidden layer and output layer. In RPNNs, the weights $\bm{w}$ and the offset bias $\beta^o$ are the only trainable parameters of the network.

Activation functions, often inspired by neuron spiking behavior (e.g., sigmoidal functions like the logistic sigmoid and hyperbolic tangent), play a crucial role in shaping the expressive power of ANNs. However, as Leshno et al. (1993) \cite{leshno1993multilayer} showed, moving beyond biological inspiration, \emph{any} non-polynomial function can achieve \emph{universal approximation}. Thus, we focus on generic non-polynomial and infinitely differentiable activation functions in the next theoretical sections. To showcase their practical potential, however, our numerical tests will utilize the familiar logistic sigmoid function.
Moreover, in the next sections, when there is no possibility of confusion, we will refer to RPNNs with $N+1$ neurons, that depends only on external weights $\bm{w}$, using the notation $f_N(\,\cdot\,;\bm{w})$, avoiding explicitly rewriting the internal fixed parameters $\bm{\alpha}$ and $\bm{\beta}$.\par

\subsection{Deep Random Projection Neural Networks}
This section describes the architecture of deep RPNNs with $L$ hidden layers and fixed internal parameters. As for the shallow case presented in Eq. \eqref{eq:RPNN}, the only trainable parameters in this setup are still the weights $\bm{w}$ and the offset $\beta^o$ between the last hidden layer and the output.\\
Consider a deep RPNN with $L$ hidden layers. Let $N_1, \ldots, N_L$ denote the number of neurons in each hidden layer. The input layer is denoted as layer 0, and the output layer is denoted as layer $L+1$. The output of neurons in layer $i$ are represented by $o_1^{(i)}, \ldots, o_{N_i}^{(i)}$. To be consistent with the notation used for the single hidden layer in Eq. \eqref{eq:RPNN} we will keep the definition of $\bm{w}$ for the external weights between the last hidden layer and the output, $\bm{\alpha}$ and $\bm{\beta}$ for internal weights and biases between input layer and first hidden layer. While stacking in the middle $L-1$ hidden layer. 
Let us denote the weights connecting the neurons from layer $i$ to layer $i+1$ by a weight matrix $W^{h}_i$, where 
the element $W^{h}_i(j, k)$ represents the weight from the $k$-th neuron in layer $i$ to the $j$-th neuron in layer $i+1$. The biases for layer $i+1$ are represented by a vector $\bm{\beta}^{(i+1)}$.
Let us denote, for simplicity, the activation functions of all the hidden neurons by $\psi(\cdot)$. 
If $L\ge2$, the output $o^{(i+1)}_{j}$ of the $j$-th neuron in layer $i+1$ is calculated as follows:
\begin{equation}
\begin{split}
    &o^{(1)}_{j} = \psi\left(\bm{\alpha}_j \cdot \bm{x} + \beta_j\right), \qquad j=1,\dots, N_1\\
    &o^{(i+1)}_{j} = \psi\left(\sum_{k=1}^{N_i} W^{h}_{i+1}(j, k) o^{(i)}_{k} + \beta_j^{(i+1)}\right),\qquad i=1,\dots,L-1, \quad j=1,\dots N_i\\
    &f_N(x;\bm{w},\beta^o,A,\bm{\beta},\bm{W}^{h}_L,\bm{B}^{h}_L) = o^{(L+1)}=\sum_{j=1}^{N_{L}} w_j o^{(L)}_{j} + \beta^{o},
\end{split}
\label{eq:deepRPNN}
\end{equation}
where $\bm{W}^{h}_L$ denotes the collection of all the internal hidden weights $W^{h}_{i+1}(j, k)$ and $\bm{B}^{h}_L$ denotes the collection of all the internal hidden biases $\beta^{(i)}_j$.
Such a configuration corresponds to highly-parametrized basis functions, that however could be equivalently denoted as $\psi_j(x)=o^{(L)}_j$ in Eq. \eqref{eq:RPNN}:
\begin{equation}
f_N(x;\bm{w},\beta^o,A,\bm{\beta},\bm{W}^{h}_L,\bm{B}^{h}_L)=\sum_{j=1}^{N_{L}} w_j \psi_j(x)+\beta^o.
\end{equation}
Indeed, regardless of the depth of the structure, the RPNN maintains only $N+1$ degrees of freedom as in Eq. \eqref{eq:RPNN}, corresponding to the size of the last hidden layer. Consequently, the findings presented in this study can be effortlessly extended to more intricate basis functions or, equivalently, to predetermined and fixed complex deep structures. However, a crucial question remains: how does stacking hidden layers with fixed internal parameters influence the network's expressiveness and learning compared to conventional deep neural networks where all parameters are trained? Exploring these questions and potential applications of deep RPNNs warrants further investigation.

\subsection{Utilizing RPNNs for Function Approximation}
When approximating a sufficiently smooth function $f:\Omega \subseteq \R\rightarrow\R$, for which we can evaluate $n+1$ points of its graph $(\bm{x}_0,y_0),(\bm{x}_1,y_1),(\bm{x}_2,y_2),\dots,(\bm{x}_n,y_n)$ such that $y_i=f(\bm{x}_i)$, the training of an RPNN reduce to the solution of a linear interpolation system of $n+1$ algebraic equation with $N+1$ unknowns $\bm{\tilde{w}}=(\beta^o,\bm{w})$:
\begin{equation}
    R\bm{\tilde{w}}=\tilde{R}\bm{w}+\beta^o=\bm{y}, \qquad R_{ij}=\psi_j(\bm{x}_i)
    \label{eq:RPNN_solve}
\end{equation}
where $\bm{y}=(y_0,y_1,\dots,y_n)$ is the vector containing the desired outputs, and the random collocation matrix $R$ has elements $R_{ij}$.\par

It has been demonstrated in \cite{ito1996nonlinearity}, that is sufficient to consider \emph{slowly increasing}\footnote{Slowly increasing (tempered) functions are function in the dual space $\mathcal{S}'(R^d)$ of Schwartz space $\mathcal{S}(R^d)$, consisting of rapidly decreasing functions.}
nonlinear \emph{plane waves}\footnote{A plane wave function is a function \( \phi: \mathbb{R}^d \rightarrow \mathbb{R} \) that can be represented as the composition of a univariate function with an affine transformation: \( \phi(\mathbf{x}) = \psi(\mathbf{a} \cdot \mathbf{x} +\mathbf{b}) \), where \( \psi: \mathbb{R} \rightarrow \mathbb{R} \) is a univariate function, and \( \mathbf{a} \) and \( \mathbf{b}\) are vectors in \( \mathbb{R}^d \).}
, such that the support of their Fourier transform $\mathcal{F}[\psi]$ has an open subset\footnote{The assumption that the support of their Fourier transform is an open subset is important. E.g. $\sin(x)$ is a slowly increasing nonpolynomial function, yet $\sin(x)=\sin(x+2\pi)$.}, as basis function and make any pair of internal weights and biases $(\bm{\alpha}_j,\beta_j)\neq \pm (\bm{\alpha}_{j'},\beta_{j'})$, $\forall j \neq j'$, to have a system of independent basis functions $\{\psi_1,\psi_2,\dots,\psi_N\}$. Note that, in various applications, the activation functions $\psi$ are often sigmoids or radial basis (Gaussian) kernels, which are slowly increasing non-polynomial functions, and their Fourier transforms have open subsets as their supports.
Let us state the following theorem here:
\begin{theorem}[Exact interpolation of RPNNs]
\label{thm:huang}
    Given an RPNN with $N$ hidden neurons as in Eq. \eqref{eq:RPNN} (or equivalently a deep RPNN as in Eq. \eqref{eq:deepRPNN}) with an infinitely differentiable, non-polynomial and slowly-increasing activation function $\psi:\R\rightarrow\R$, such that the support of its Fourier transform is an open subset. Let internal weights $A$ and biases $\bm{\beta}$ be randomly chosen according to any continuous probability distribution. Then, for $N+1$ arbitrary input-output samples $(\bm{x}_i,y_i)$, with distinct inputs $\bm{x}_i\neq \bm{x}_{i'}, \, \forall i\neq i'$, there exist a unique choice of external weights $\bm{w}$ and offset $\beta^o$ that ensure with probability 1:
\begin{equation}
    \mathbb{P}(|f_N(\bm{x}_i;\bm{w},\beta^o,A,\bm{\beta})-y_i|=0)=1, \qquad \forall i=1,\dots,n+1
\end{equation}
\end{theorem}
For the proof of the above theorem see in Appendix \ref{sec:proof_interpolation}.

However, even if the matrix $R$ in Eq. \eqref{eq:RPNN_solve} is invertible under the condition of Theorem \ref{thm:huang}, for many random choices of internal weights $\bm{\alpha},\bm{\beta}$ and collocation points $\bm{x}$, in practice the random collocation matrix $R$ numerically tends to be ill-conditioned and close to singular. Therefore, in practice, it is often suggested to solve Eq. \eqref{eq:RPNN_solve} via a Moore-Penrose pseudo-inverse, that can be computed by a truncated singular value decomposition (tSVD). This involves computing the decomposition and the regularized pseudo-inverse $R^{\dagger}$, as follows:
\begin{equation}
    R=U\Sigma V^T=[U_q \quad \tilde{U}]
    \begin{bmatrix}
        \Sigma_q & 0\\
        0 & \tilde{\Sigma}
    \end{bmatrix}
     [V_q \quad \tilde{V}]^T
    , \qquad R^{\dagger}=V_q\Sigma_q^{-1}U_q^T,
    \label{eq:pseudo_inverse}
\end{equation}
where the matrices $U=[U_q \quad \tilde{U}]\in \R^{n+1\times n+1}$ and $V=[Vq \quad \tilde{V}] \in \R^{N+1\times n+1}$ are orthogonal and $\Sigma \in \R^{n+1\times N+1}$ is a diagonal matrix containing the singular values $\sigma_i=\Sigma_{(i,i)}$. Here, we select the $q$ largest singular values exceeding a specified tolerance $0<\epsilon\ll 1$, i.e., $\sigma_1,\dots,\sigma_q>\epsilon$, effectively filtering out insignificant contributions and improving numerical stability.  
Note that the solution of Eq. \eqref{eq:RPNN_solve} with the Moore-Penrose pseudo-inverse correspond to the solution of the following regularized least square problem:
\begin{equation}
    \arg\min_{\bm{\tilde{w}} \in \R^{N+1}} \|R\bm{\tilde{w}}-\bm{y}\|_2^2+\epsilon\|\bm{\tilde{w}}\|_2^2.
    \label{eq:minimum_norm}
\end{equation}
Alternatively, a more robust method that further enhances numerical stability, involves utilizing a rank-revealing QR decomposition with column-pivoting:
\begin{equation}
    R\, P= [Q_1 \quad Q_2]\begin{bmatrix}
        T\\
        0
    \end{bmatrix}, 
\end{equation}
where the matrix $Q=[Q_1 \quad Q_2] \in \R^{n+1\times n+1}$ is orthogonal, $T \in \R^{N+1 \times N+1}$ is an upper triangular square matrix and the matrix $P \in \R^{n+1 \times n+1}$ is an orthogonal permutation of the columns. The key advantage of the column permutations lies in its ability to automatically identify and discard small values that contribute to instability. Indeed, in case of ill-conditioned (rank-deficient) system we have that effectively the column of the matrix $Q$ do not span the same space as the column of the matrix $R$. As a result the matrix $T$ is not full upper triangular but we have:
\begin{equation}
    R=[Q_1 \quad Q_2]
    \begin{bmatrix}
        T_{11} & T_{12}\\
        0 & 0
    \end{bmatrix}
    P^T,
    \label{eq:QR1}
\end{equation}
where, if $rank(R)=r<N+1$, the matrix $T_{11} \in \R^{r \times r}$ is effectively upper triangular and $T_{12} \in \R^{r\times (N+1-r)}$ are the remaining columns. Note that numerically, one selects a tolerance $0<\epsilon<<1$ to estimate the rank $r$ of the matrix $R$ and set values of $T$ below the threshold to zero. 
Hence, based on \eqref{eq:QR1}, we can solve $R\bm{\tilde{w}}=\bm{y}$, by first setting $P^T\bm{\tilde{w}}=[\bm{z}_1 \, \bm{z}_2]^T$ and solve for $\bm{z}_1 \in \R^r$ the following system:
\begin{equation}
    T_{11}\bm{z}_1=Q_1^T\bm{y}-T_{12}\bm{z}_2.
    \label{eq:lastQR}
\end{equation}
Note that casting the back-substitution algorithm for the matrix $T_{11}$ in Eq. \eqref{eq:lastQR}, we can efficiently solve the system. In order to also obtain the \emph{minimum-norm} solution, one can additionally cast the Complete Orthogonal Decomposition (COD), by also computing the QR decomposition of the transposed non-zero elements in $T$:
\begin{equation}
    \begin{bmatrix}
        T_{11}^T \\
        T_{22}^T
    \end{bmatrix}
    =V
    \begin{bmatrix}
        \tilde{T}_{11}^T\\
        0
    \end{bmatrix}.
\end{equation}
Finally, by setting $S=PV$, one obtains:
\begin{equation}
    R=[Q_1 \quad Q_2]\begin{bmatrix}
        \tilde{T}_{11} & 0\\
        0 & 0
    \end{bmatrix}
    (S)^T,
    \label{eq:COD}
\end{equation}
where $\tilde{T}_{11}$ is a lower triangular matrix of size $r \times r$. Finally, the least-square minimum norm solution, as in Eq. \eqref{eq:minimum_norm}. is given by:
\begin{equation}
    \bm{\tilde{w}}=S \cdot (\tilde{T}_{11}^{-1}(Q_1^T \bm{y}))
\end{equation} 
where the inversion of $\tilde{T}_{11}$ is computed casting the forward substitution algorithm that is numerically stable.

\section{Best Approximation with RPNN}
\label{sec:best}
Let us consider the case in which the input is one-dimensional $x \in [a,b] \subset \R$. The Eq. \eqref{eq:RPNN} can be restated as:
\begin{equation}
f_N(x;\bm{w},\beta^o,\bm{\alpha},\bm{\beta})=\sum_{j=1}^N w_j \psi (\alpha_j \cdot x+ \beta_j)+\beta^o=\sum_{j=1}^N w_j \psi_j (x)+\beta^o,
\label{eq:RPNN_N+1_1d}
\end{equation}
where now we represent the internal weights by the column vector $\bm{\alpha}=(\alpha_1,\dots,\alpha_N) \in \R^{N\times1}$.
Here, we define the concept of RPNN of best approximation and provide proof about its existence and uniqueness.
Let us first trivially note that if we consider an RPNN with independent fixed basis function, we have a linear space:
\begin{theorem}
Let $\psi$ be a slowly increasing (or infinitely differentiable) activation function.
    The set $\mathcal{M}_{\! (N,\bm{\alpha},\bm{\beta})}^{[a,b]}=\{f_N \in C[a,b] : f_N(x;\bm{w},\beta^o,\bm{\alpha},\bm{\beta})=\sum_{j=1}^N w_j \psi (\alpha_j x+ \beta_j)+\beta^o, \text{ with } (\alpha_j,\beta_j)\neq \pm(\alpha_{j'},\beta_{j'}) \text{ and } \bm{w} \in \R^{N}\}$ of RPNN is a vector space of dimension $N+1$ on $\R$ with usual vector sum and scalar vector product.
\begin{proof}
For $(\alpha_j,\beta_j)\neq\pm(\alpha_i,\beta_i)$ the set of basis function is independent for \cite{ito1996nonlinearity}. Then clearly $f_N(\, \cdot\,;\bm{tilde{w}})\equiv0$, with $\bm{\tilde{w}}=(\beta^o,\bm{w})=\bm{0}$, is in $\mathcal{M}_{\! (N,\bm{\alpha},\bm{\beta})}^{[a,b]}$ and since if we have $f_N(\,\cdot\, ;\bm{w}_1,\beta^o_1),f_N(\, \cdot \, ;\bm{w}_2,\beta^o_2)\in \mathcal{M}_{\!(N,\bm{\alpha},\bm{\beta})}^{[a,b]}$, based on \eqref{eq:RPNN_N+1_1d}, also we obtain $f_N(x;\gamma_1 \bm{w}_1 + \gamma_2\bm{w}_2,\gamma_1\beta_1^o+\gamma_2\beta_2^o), \, \forall \gamma_1,\gamma_2$.
Finally, trivially, there exist an isomorphism between the set of external weights $\bm{\tilde{w}}$ and the usual vector space on $\R^{N+1}$.
\end{proof}    
\end{theorem}
Let's also observe that the differentiability of the function of $\mathcal{M}_{\! (N,\bm{\alpha},\bm{\beta})}^{[a,b]}$, depends on the differentiability of the activation functions. Thus:
\begin{lemma}
    If the activation function $\psi \in C^{\nu}(\R)$, then the space $\mathcal{M}_{\! (N,\bm{\alpha},\bm{\beta})}^{[a,b]}$ is a vector subspace of $C^{\nu}([a,b])$.
\end{lemma}

Now, we proceed by considering a norm, such as the $L^p$-norm with $p=1,2,\infty$, denoted also as $\|\cdot\|_{\infty}$, $\|\cdot\|_2$, or $\|\cdot\|_1$. Since $C[a,b]$ equipped with $L^p$ norms is a Banach space, it follows that $\mathcal{M}_{! (N,\bm{\alpha},\bm{\beta})}^{[a,b]}$ is also a Banach space.

\begin{definition}
    We define the RPNN of best approximation in norm $L^p$ over $\mathcal{M}_{\bm{\alpha},\bm{\beta}}$ as the RPNN which solve
\begin{equation}
\| f - f_N\|_{p} = \min_{g_N \in \mathcal{M}_{\! (N,\bm{\alpha},\bm{\beta})}^{[a,b]}} \| f-g_N\|_{p}.
\label{eq:best_Lp}
\end{equation}
\end{definition}

\subsection{Existence}
Since $\mathcal{M}_{\! (N,\bm{\alpha},\bm{\beta})}^{[a,b]}$ is a finite dimensional Banach space, and the function $\|f-\cdot\|_{p}$ is continuous, for the Stone–Weierstrass approximation theorem, there exists a ANN as defined above of best $L^p$ approximation in $\mathcal{M}_{\! (N,\bm{\alpha},\bm{\beta})}^{[a,b]}$. Indeed, the zero function, described by the vector of parameters $\bm{0}$, is in $\mathcal{M}_{\! (N,\bm{\alpha},\bm{\beta})}^{[a,b]}$, which approximate $f$ with error $\|f\|_{p}$. So the best approximation should be in the closed ball of function $g_N \in \mathcal{M}_{\! (N,\bm{\alpha},\bm{\beta})}^{[a,b]}$ such that $\|f-g_N\|_{p}\le\|f\|_{p}$, which is closed and limited, so it is a compact set. Therefore a global minimum in $\mathcal{M}_{\! (N,\bm{\alpha},\bm{\beta})}^{[a,b]}$ exists. \par

\subsection{Uniqueness}
We have shown that $\mathcal{M}_{\! (N,\bm{\alpha},\bm{\beta})}^{[a,b]}$ forms a vectors space, implying it is also a convex set. For illustration, consider two vectors of weights $\bm{\tilde{w}}_0=(\beta^o_0,\bm{w}_0)$ and $\bm{\tilde{w}}_1=(\beta^o_1,\bm{w}_1)$ and a point $\bm{\tilde{w}}_{\gamma}$ on the segment, i.e.: 
\begin{equation}
    \bm{\tilde{w}}_{\gamma}=\gamma \bm{\tilde{w}}_1 + (1-\gamma) \bm{\tilde{w}}_0, \qquad \gamma \in [0,1].
    \label{eq:segment_gamma_w}
\end{equation}
The vector $\bm{\tilde{w}}_{\gamma}$ still represents the external weights of an RPNN $f_N(\, \cdot \, ;\bm{\tilde{w}}_{\gamma})$ in $\mathcal{M}_{\! (N,\bm{\alpha},\bm{\beta})}^{[a,b]}$. Hence, we can say that, there exists either a unique best $L^p$ approximation within the space $\mathcal{M}_{\! (N,\bm{\alpha},\bm{\beta})}^{[a,b]}$ or infinitely many equally good approximations, all corresponding to the points along the segment.
Let's assume $\bm{\tilde{w}}_0$ and $\bm{\tilde{w}}_1$ are two solutions of best $L^p$ approximation of a function $f$, thus $\|f-f_N(\,\cdot\, ;\bm{\tilde{w}}_0)\|_{p}=\|f-f_N(\,\cdot\, ;\bm{\tilde{w}}_1)\|_{p}=r$. Then for $\bm{\tilde{w}}_{\gamma}$ as in Eq. \eqref{eq:segment_gamma_w}:
\begin{equation}
\begin{split}
    &\|f-f_N(\bm{\tilde{w}}_{\gamma})\|_{p} = \|\gamma (f-f_N(\,\cdot\, ;\bm{\tilde{w}}_1))+(1-\gamma) (f-f_N(\,\cdot\, ;\bm{\tilde{w}}_0))\|_{p} \le\\
    &\le \gamma \|f-f_N(\,\cdot\, ;\bm{\tilde{w}}_1)\|_{p}+(1-\gamma)\|f-f_N(\,\cdot\, ;\bm{\tilde{w}}_0)\|_{p} = \gamma r + (1-\gamma) r = r
\end{split}
\end{equation}
Hence, to maintain consistency with the best approximations defined by $\bm{\tilde{w}}_0$ and $\bm{\tilde{w}}_1$, we require necessarily the equality $\|f-f_N(\,\cdot\,;\bm{\tilde{w}}_{\gamma})\|_{\infty}=r$ to hold $\forall \gamma \in [0,1]$. This implies that all $\bm{\tilde{w}}_{\gamma}$ define alternative RPNNs with the same maximum error as the original best approximations.
If we consider $L^p$ norms, with $1<p<\infty$, the space is also strictly convex\footnote{A normed linear space $(X,\|\cdot\|)$ is called strictly convex if for $u,v \in X$, $\|u\|\le r\|$ and $\|v\|\le r$, $\|u+v\|<2r$ unless $u=v$.}, which imply that necessarily $\bm{\tilde{w}}_1=\bm{\tilde{w}}_2$. Otherwise, if they are distinct then, by strict convexity:
\begin{equation}
    \|f-f_N(\,\cdot\, ;\bm{\tilde{w}}_0)+f-f_N(\,\cdot\, ;\bm{\tilde{w}}_1)\|_p<2r
\end{equation}
or, equivalently,
\begin{equation}
    \big\|f-\frac{f_N(\,\cdot\, ;\bm{\tilde{w}}_0)+f_N(\,\cdot\, ;\bm{\tilde{w}}_1)}{2}\big\|_p<r
\end{equation}
thus we obtain a better approximation with $\bm{\tilde{w}}_{\gamma}=\bm{\tilde{w}}_{1/2}$
\begin{equation}
    \|f-f_N(\,\cdot\,;\bm{\tilde{w}}_{1/2})\|_p<r,
\end{equation}
that is a contradiction. Therefore, we can state the following theorem.
\begin{theorem}
    The best $L^p$ approximation problem with RPNNs in Eq. \eqref{eq:best_Lp} has a unique solution in the space $\mathcal{M}_{\! (N,\bm{\alpha},\bm{\beta})}^{[a,b]}$ when $1<p<\infty$.
\end{theorem}

\section{Main result: Exponential Convergence}
\label{sec:main}
In this section, we will prove that by considering activation functions that are infinitely differentiable, then the RPNNs of best $L^p$ approximation can exhibit exponential convergence towards approximating an infinitely differentiable function.
Let us assume we want to approximate the function $f:\R\rightarrow\R$. Then we state the following theorem 
\begin{theorem}
    The RPNN of best $L^p$ approximation in $\mathcal{M}_{\! (N,\bm{\alpha},\bm{\beta})}^{[a,b]}$ equipped with an infinitely differentiable non-polynomial activation function converge exponentially fast when approximating an infinitely differentiable function $f:\R\rightarrow\R$.
    \label{thm:mainthm}
\end{theorem}
Our goal is to demonstrate the existence of a choice of external weights and biases that exponentially enhance the convergence rate of an RPNN approximation. Subsequently, the unique RPNN configured for the best approximation is guaranteed to converge at least as rapidly. Specifically, we establish the following Lemma:
\begin{lemma}
There exist a choice of weights $\bm{\tilde{w}}=(\beta^o,\bm{w})$ that makes an RPNN in $\mathcal{M}_{\! (N,\alpha, \beta)}^{[a,b]}$ with infinitely differentiable non-polynomial activation functions, converging exponentially fast in norm $L^p$ when approximating an infinitely differentiable function $f:\R\rightarrow\R$.
\begin{proof}
 Let's call $\mathcal{P}_n$ the polynomial of best $L^p$ approximation of degree $n$ of $f$ in the interval $[-1,1]$ w.r.t. the norm $\|\cdot\|_p$ with $1<p\le \infty$:
\begin{equation}
f(x) \simeq \mathcal{P}_n(x)=\sum_{k=0}^n a_k x^k.
\label{eq:p_poly}
\end{equation}
 Let's call $f_N(\,\cdot\,;\bm{\tilde{w}}) \in \mathcal{M}_{\! (N,\alpha, \beta)}^{[a,b]}$ an RPNN approximation of $f$ that we are seeking.
\begin{equation}
f(x) \simeq f_N(x;\bm{\tilde{w}})= \sum_{j=1}^n w_j \psi (\alpha_j \cdot x +\beta_j)+\beta^o.
\end{equation}
where $\psi$ is the activation function which is a non-polynomial and infinitely differentiable.\\
Let us also define $c_j$ the \emph{centers} of the wave plane activation function such that $c_j=-\frac{\beta_j}{\alpha_j}$ (see e.g. \cite{fabiani2021numerical,calabro2021extreme}), or equivalently, $\beta_j=-\alpha_j \cdot c_j$:
\begin{equation}
f(x) \simeq f_N(x;\bm{\tilde{w}})= \sum_{j=1}^N w_j \psi (\alpha_j \cdot (x -c_j))+\beta^o.
\label{eq:first_centers}
\end{equation}
Let us consider also the best $L^p$ approximation polynomial $\mathcal{Q}_n$ of degree $n$ of the activation function $\psi$, in the interval $I_{\alpha,\beta}=[\min(\alpha_j \cdot x +\beta_j),\max(\alpha_j \cdot x +\beta_j)]$:
\begin{equation}
\psi(x) \simeq \mathcal{Q}_n(x)= \sum_{k=0}^n b_k x^k.
\end{equation}
Note that since $\psi$ is a \emph{non-polynomial} function, it cannot be \emph{exactly} (with zero residual) represented by any polynomial $\mathcal{Q}_n$ of finite degree. Therefore, in what follows we can consider arbitrary large $n$, and 
a corresponding \emph{non-zero} coefficient $b_n$.
Also note that the interval $I_{\alpha,\beta}$ is considered, to maintain accurate approximation of $\psi_j$ by using the transformed polynomial $\mathcal{Q}_n(\alpha_j x +\beta_j)$.

We can then approximate the neural network $f_N$ with the polynomial $\mathcal{G}_{N,n}$, given by:
\begin{equation}
f_N(x;\bm{w}) \simeq \mathcal{G}_{N,n}(x)= \sum_{j=1}^N w_j \sum_{k=0}^n b_k \alpha_j^k (x-c_j)^k+\beta^o.
\end{equation}
We can rewrite the polynomial $\mathcal{G}_{N,n}$ using Newton's binomial expansion:
\begin{equation}
\mathcal{G}_{N,n}(x)= \sum_{j=1}^N w_j \sum_{k=0}^n b_k \alpha_j^k \sum_{s=0}^k \binom{k}{s} (-c_j)^{n-s}x^s+\beta^o.
\label{eq:g_expansion}
\end{equation}
At this point, we want to prove that we can find a vector of coefficient $\bm{\tilde{w}}$ such that we can have $\mathcal{G}_{N,n}=\mathcal{P}_n$, thus we have to equate the coefficients of order $k$ in Eq. \eqref{eq:g_expansion} and $\eqref{eq:p_poly}$, thus resulting in a system of $n$ equations in $N$ unknowns:
\begin{equation}
a_k= \sum_{j=1}^N w_j (-c_j)^{n-k} \sum_{s=k}^n \binom{s}{k} b_s \alpha_j^s, \qquad k=1,\dots,n,
\label{eq:coeff_gNn}
\end{equation}
while trivially equate the two offsets $\beta^o=a_0$. Eq. \eqref{eq:coeff_gNn} can be written in matrix form as
\begin{equation}
\bm{a}=\bm{w}\cdot M,
\end{equation}
where $\bm{a}$ is the vector containing the coefficient $a_k$ and $M$ is the matrix with elements $M_{k,j}$ that reads:
\begin{equation}
M_{k,j}= (-c_j)^{n-k} \sum_{s=k}^n \binom{s}{k}b_s \alpha_j^s.
\label{eq:A_matrix}
\end{equation}
If we take $N=n$ and we prove that the matrix $M$ in Eq. \eqref{eq:A_matrix} is invertible, with inverse $M^{-1}$, then there exist a unique choice of weights $\bm{w}=\bm{a}\cdot M^{-1}$ that make $\mathcal{G}_{N,N}=\mathcal{P}_N$.
Note that as proved in \cite{ito1996nonlinearity}, the slowly increasing activation functions are independent if $(\alpha_j,\beta_j) \neq
\pm(\alpha_{j'},\beta_{j'})$.
Thus, we may allow both $\alpha_j$ and $c_j$ to be different for $j\neq j'$.
Let us consider a fixed index $k=1,\dots,N$ in Eq. \eqref{eq:A_matrix}, then we have a polynomial in the two variables $\alpha, c$ of type:
\begin{equation}
\tau_N c^{N-k}\alpha_k+\tau_{N-1} c^{N-k}\alpha_{k+1}...+\tau_{N-k}c^{N-k}\alpha_{N},
\end{equation}
where $\tau$s are the coefficients in Eq. \eqref{eq:A_matrix}. Different index $k'\neq k$ corresponds to polynomials of different order, which, as can be noted do not have any monomials in common. Indeed $c^{N-k}\neq c^{N-k'}$, thus they are a set of independent polynomials. Therefore, $M$ in Eq. \eqref{eq:A_matrix} is invertible.\par

Finally, when approximating $f$ with an RPNN $f_{N}(\,\cdot\,;\bm{\tilde{w}})$, where the weights $\bm{\tilde{w}}$ make $f_N$ coincide with the polynomial $\mathcal{G}_{N,N}$ constructed as above, we can prove that $f_N(\,\cdot\,;\bm{\tilde{w}})$ exponentially convergence to $f$, since:
\begin{equation}
\begin{split}
    &\|f-f_N\|_{p}=\|f-\mathcal{P}_N+\mathcal{P}_N-\mathcal{G}_{N,N}+\mathcal{G}_{N,N}-f_{N}\|_{p}\le\\
&\le \|f-\mathcal{P}_N\|_{p}+\|\mathcal{P}_N-\mathcal{G}_{N,N}\|_{p}+\|\mathcal{G}_{N,N}-f_{N}\|_{p}.
\end{split}
\label{eq:convergence_theorem}
\end{equation}
But since $\mathcal{P}_N$ is the best approximation polynomial, then $\|f-\mathcal{P}_N\|_{p}$ converge exponentially, as well as $\|\mathcal{G}_{N,N}-f_{N}\|_{p}$ converge exponentially because we have used the best approximation polynomial $\mathcal{Q}_N$ of the activation function $\psi$. Finally $\|\mathcal{P}_N-\mathcal{G}_{N,N}\|_{p}=0$, because of the invertibility of the matrix $M$ in \eqref{eq:A_matrix}.
    \end{proof}
\end{lemma}

Please note that the above constructive proof has shown that one has to invert a matrix $M$ in Eq. \eqref{eq:A_matrix} which shares some similarities with the Vandermonde matrix, especially as $N$ becomes large or as the degrees of the polynomials increase, therefore it may be an ill-conditioned matrix.

In order to evaluate the upper bound of the convergence as we have determined in Eq. \eqref{eq:convergence_theorem}. Let us restate a classical result about the upper bounds of convergence of Legendre polynomials (see Chapter 2 in Gaier (1987) \cite{gaier1987lectures}).
\begin{theorem}
\label{threfeten}
Let a function $f$ analytic in $[-1,1]$, that is analytically continuable to the open Bernstein Ellipse $\mathcal{E}_{\rho}$, with $1\le\rho\le \infty$, in the complex plane, given by:
\begin{equation}
    \mathcal{E}_{\rho}:=\biggl\{ z \in \C \, \biggl| \, z=\frac{u+u^-1}{2}, u=\rho e^{i\theta}, 0\le \theta\le 2\pi \biggr\}
\end{equation}
and let ${\mathcal{P}_N}$ be the Legendre interpolants to $f$ in Legendre grid of $N+1$ points ${x_N} \in [-1,1]$. Then the error satisfies:
\begin{equation}
    \lim_{N\rightarrow \infty}\|f-\mathcal{P}_N\|_{p}=\rho^{-N}.
    \label{eq:convergence_legendre}
\end{equation}
\end{theorem}
a polynomial interpolant satysfing Eq. \eqref{eq:convergence_legendre} is said to be \emph{maximally convergent}. Other examples of such polynomials are interpolants in Chebyshev, or Gauss-Jacobi grids. The effective convergence rates of these polynomials may differ because of some algebraic factors. In particular. the value of $\rho$ depends on the smoothness of the function $f$ and on the domain interval $[a,b]$, that need to be rescaled to $[-1,1]$, by considering the Bernstein Ellipse $\mathcal{E}_{\rho}$ associated with $\tilde{f}(x)=f\big(\frac{2x -a-b}{b-a}\big):[-1,1]\rightarrow \R$.

Now, we can note that in Eq. \eqref{eq:convergence_theorem}, $\|f-\mathcal{P}_N\|_p$ in the interval $[a,b]$ converge as $O(\rho(f,[a,b])^{-N})$, where $\rho(f,[a,b])>1$ is the constant associate to the maximum Bernstein ellipse $\mathcal{E}_{\rho}$ in which $\tilde{f}$ is analytically continuable.  While the convergence of $\|\mathcal{G}_{N,N}-f_{N}\|_p$ is given by $N$ times the convergence of $\|\mathcal{Q}_N-\psi\|$, thus $\|\mathcal{G}_{N,N}-f_{N}\|_{p}$ is of order $O(N \cdot \rho(\psi,I_{\alpha,\beta})^{-N})$. \par

Finally, we emphasize that since the Theorem \ref{thm:mainthm} hinges on the ability of RPNNs to accurately reproduce polynomials, as described in the upper bound in Eq. \eqref{eq:convergence_theorem} when approximating less smooth functions, with a maximum order of differentiability $\nu\ge 0$, this sets a limit to the achievable convergence rate, reducing it to $O(N^{-\nu})$.

\section{On the selection of the internal parameters}
\label{sec:selection}
In this section, we delve into the \emph{a priori} selection of internal weights and biases for the RPNN approximator. 
Despite the theoretical assurance that any random parameter selection should lead to exponential convergence, as proven in Theorem \ref{thm:mainthm}, practical considerations necessitate a judicious choice to mitigate the ill-conditioning of matrix $R$ in Eq. \eqref{eq:RPNN_solve}.

Moving beyond a \textit{naive random generation} \cite{huang2006extreme} of internal parameter, we therefore explore two alternative strategies for selection, namely:

\begin{itemize}
    \item \textit{Parsimonious Function-Agnostic Selection:} Parameters dependent solely on the domain of data of the sought function.
    \item \textit{Function-Informed Selection:} Addressing the specific shape of the function in consideration.
\end{itemize}

\subsection{Naive random generation}
\label{sec:random_selection}
In a naive random generation approach for internal parameters and biases of an RPNN \cite{huang2006extreme}, a common practice involves the normalization of input $\bm{x} \in [a,b]$ and output data $\bm{y} \in [c,d]$ within the domain $[-1,1]$. Thus, employing these transformation ${x}\rightarrow \tilde{{x}}$ and ${y}\rightarrow \tilde{{y}}$:
\begin{equation}
\begin{split}
    \tilde{x}=\frac{2{x}-a-b}{b-a} \in [-1,1], \qquad \tilde{y}=\frac{2{y}-c-d}{d-c} \in [-1,1].
\end{split}
\label{eq:normalization}
\end{equation}
Subsequently, weights $\bm{\alpha}=(\alpha_1,\dots,\alpha_N)$ are uniformly and randomly selected from the range $[-1,1]$, and biases $\bm{\beta}=(\beta_1,\dots,\beta_N)$ are similarly chosen within the domain $[-1,1]$.
Despite its simplicity, this method has proven effective in certain scenarios, making it a straightforward yet powerful approach for initializing RPNNs \cite{huang2006extreme}. However, as numerical results will demonstrate, this choice deviates considerably from optimal configurations.

\subsection{Parsimonious function-agnostic selection}
\label{sec:random_agnostic_selection}
As proposed in previous works \cite{fabiani2023parsimonious,fabiani2021numerical,calabro2021extreme}, a fundamental strategy for obtaining effective basis functions involves ensuring that the activation functions $\psi_j$ have centers $c_j=-\beta_j/\alpha_j$ (as defined in Eq. \eqref{eq:first_centers}) within the domain $[a,b]$ of interest.
This choice depends on the specific activation function, and for simplicity, we focus on common sigmoid-like functions, such as the logistic sigmoid, given by:
\begin{equation}
    \psi_j(x)=\frac{1}{1+\text{exp}(-\alpha_j x -\beta_j)}
    \label{eq:logistic_sigmoid}
\end{equation}
The parameter $\alpha_j$ plays a crucial role in determining the sharpness of the sigmoid. For instance, setting $\alpha_j=0$ results in a constant function, while small values of $|\alpha_j|<1$ yield almost-linear functions. On the other hand, for significantly large values of $|\alpha_j|>>1$, an almost-Heaviside function is obtained.
As done in \cite{fabiani2021numerical}, we choose $\bm{\alpha}$ to be a vector of i.i.d  random uniformly distributed values in a range that depends on the number of neurons $N$; if the domain of interest is $[a,b]$, we propose to set:
\begin{equation}
    \alpha_j \sim \mathcal{U}\biggl[-\frac{(400+9N)}{10(b-a)}),\frac{(400+9N)}{10(b-a)}\biggr], \qquad j=1,\dots, N
\end{equation}
where $\mathcal{U}$ denotes the uniform distribution.
\par
The bias parameter $\beta_j$ influences the position of the center $c_j$ of the activation function.
One can either decide these centers beforehand (e.g. equally spaced) or randomly sample them within the domain of interest and subsequently determine the corresponding bias values using the following relation $\beta_j=-\frac{c_j}{\alpha_j}$.

\subsection{Function-informed selection}
\label{sec:random_informed_selection}
In contrast to the just described \emph{function-agnostic} selection, here we propose a \emph{function-informed} selection for internal parameters and biases of an RPNN, that involves a more systematic approach. Here, we employ equally spaced centers in the input domain and leverage a centered finite difference (FD) scheme to efficiently compute the first derivative of the input function. Utilizing this derivative information, we try to set accordingly the $\alpha_j$ parameters for each neuron.
This method ensures a more tailored initialization based on the specific characteristics of the input function. Here below, an example for the specific case of a logistic sigmoid in Eq. \eqref{eq:logistic_sigmoid}, which have analytical derivative:
\begin{equation}
\label{eq:der_logistic_sigmoid}
    \frac{d\psi_j(x)}{dx}=\frac{\alpha_j\exp(\alpha_j x + \beta_j)}{(1+\exp(\alpha_j x+\beta_j)^2}=\frac{\alpha_j\exp\big(\alpha_j (x -c_j)\big)}{\big(1+\exp(\alpha_j (x-c_j)\big)^2}
\end{equation}
where $c_j$ are the centers defined as $c_j=-\alpha_j\beta_j$. First as for the naive generation, we normalize the input and the output data as in Eq. \eqref{eq:normalization}. Hence we consider the rescaled function \(\tilde{f}\).
Then we set the centers $\tilde{c}_j$ equally spaced in $[-1,1]$. Then (b) we compute the approximated derivatives of $\tilde{f}'$ of the function $\tilde{f}$ at the centers $\tilde{c}_j$ with a centered FD scheme. By observing that the derivative of the logistic sigmoid in Eq. \eqref{eq:der_logistic_sigmoid} evaluated at the center $\tilde{c}_j=-\beta_j/\alpha_j$ is:
    \begin{equation}
        \frac{d \psi_j(c_j)}{dx}=\frac{\alpha_j}{4},
    \end{equation} 
Finally, (c) we locally assume that:
    \begin{equation}
       \frac{d \psi_j(c_j)}{dx}=\tilde{f}'(c_j).
        \label{eq:assumption_derivative}
    \end{equation}
The above assumption is not generally correct as the contribution for the derivative of the RPNN $f_N(\cdot, \bm{w})$ is not only given by the single activation function $\psi_j$, but is also influenced by the corresponding weight $w_j$ and possibly by all the other activations functions. However, in the case of a sharp gradient, a sharp sigmoid is suitable and the main variation (the highest derivative value) is localized at $c_j$, while far from $c_j$ the sigmoids becomes rather constants. Therefore, we set ultimately and heuristically:
\begin{equation}
    \alpha_j =\gamma \frac{f(c_{j+1})-f(c_{j-1})}{ \Delta x}+\varepsilon_j,
\end{equation}
where $\Delta x$ is the distance between two consecutive centers, the proportionality constant is set $\gamma=3/2$ and $\varepsilon_j$ is a random variable uniformly distributed, accounting for the inexactness of the assumption in \eqref{eq:assumption_derivative}, as:
\begin{equation}
    \varepsilon_j \sim \mathcal{U}\biggl[-\frac{(400+9N)}{100(b-a)}),\frac{(400+9N)}{100(b-a)}\biggr], \qquad j=1,\dots, N.
\end{equation}

\section{Numerical examples}
\label{sec:numerical}
In this section, we conduct numerical tests of the proposed RPNN-based best $L^2$ approximation method as in Eq. \eqref{eq:best_Lp}, using five benchmark functions. The training of the RPNNs is performed by employing the Complete Orthogonal Decomposition (COD) as explained in Eq. \eqref{eq:COD}, where we set the tolerance $\epsilon$ for the rank estimation to:
\[\epsilon=n\cdot \texttt{eps}(\|R\|_2)/1000,\]
where $n$ is the number of data points, $R$ is the matrix in \eqref{eq:RPNN_solve} and \texttt{eps} is the built-in MATLAB function returning the floating-point relative accuracy (e.g., \texttt{eps}(1.0)=2.2204E$-$16).
Additionally, in Appendix \ref{sec:SVDvsCOD} we provide a comparison of the use of the SVD-based and the COD-based least-square minimum norm solutions for different values of the tolerance $\epsilon$.

We compare RPNNs with classical numerical analysis methods, including the Legendre Polynomials, implemented through a Lagrange barycentric interpolation formula, and the Cubic spline, implemented with the \texttt{spline} built-in function of \texttt{MATLAB}.
For each benchmark, since the selection of the RPNN's internal parameter is random, we performed $100$ different Monte-Carlo selections drawn by their respective distributions, and we report the mean approximation accuracy. We highlight that, even if RPNNs are basically random algorithms, the results confirm their robustness.
For the implementation of different strategies for the random selections of the internal parameters of the RPNNs, see Section \ref{sec:selection}.
In particular in the following for all the tests, we have compared the use of (a) naive random generated RPNNs; (b) function-agnostic RPNNs and (c) function-informed RPNNs, without employing any further training of the basis functions.
Additionally, for comparison purposes we also compared with a standard single hidden layer feedforward neural network (FNN), \emph{fully trainable}, with $N$ neurons in the hidden layer and logistic sigmoid activation functions, as implemented by the \texttt{MATLAB} deep learning toolbox (e.g., the function \texttt{net=feedforwardnet(N)}).
Note that a FNN has more free parameters than a similar RPNN ($3N+1$ versus $N+1$). However, we think it is fair to compare the two networks, that share the same architecture, just in terms of $N$. Note that in the comparison $N$ is also the order of the Legendre polynomials and number of interpolation nodes used for cubic spline.

For the training process of the FNNs, we utilized the Levenberg-Marquardt algorithm, as implemented in \texttt{MATLAB} by the option \texttt{net.trainFcn=`trainlm'}, with $1000$ as maximum number of epochs and $1e-10$ as minimum gradient of the loss function.
In the following case studies, \texttt{trainlm} was able to obtain lower mean square errors than any of the other algorithms we tested. Moreover, the training process is influenced by the first random initialization of the weights, therefore, we repeated $10$ times the training starting from different random initializations, and we report the best performing FNN from these runs.
For each number of neurons $N$, both FNN and RPNNs were trained on $n=5N$ equally spaced data points within the domain of interest (training set).
\begin{figure}[ht!]
    \centering
    \subfigure[references]{\includegraphics[width=0.31\textwidth]{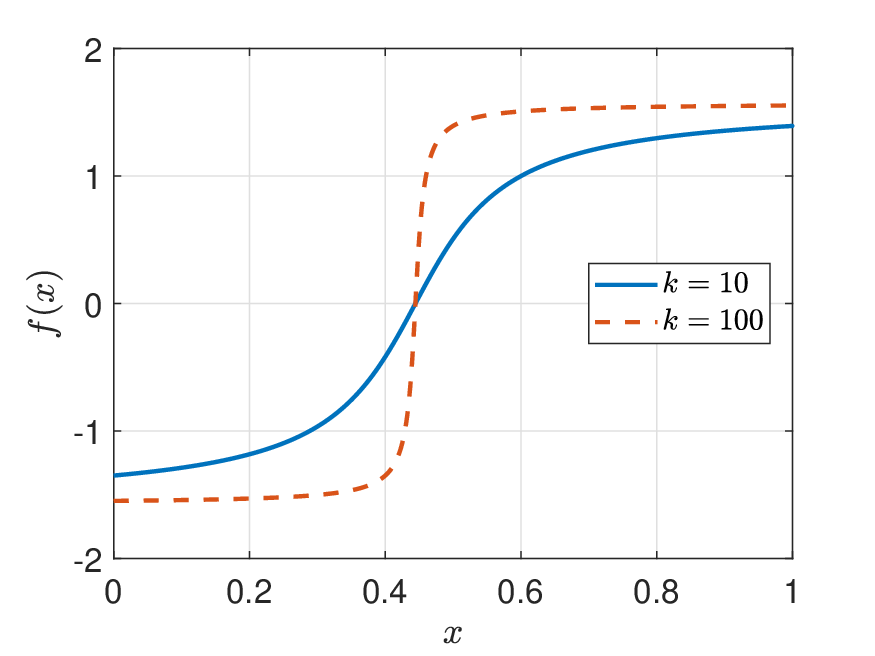}}
    \subfigure[results $k=10$]{
    \includegraphics[width=0.31\textwidth]{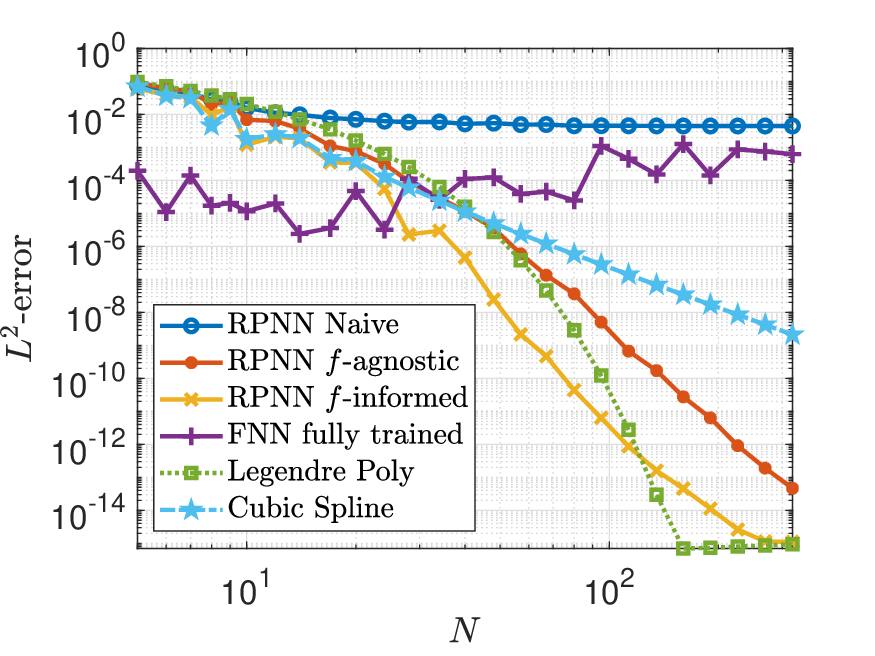}
    }
    \subfigure[results $k=100$]{\includegraphics[width=0.31\textwidth]{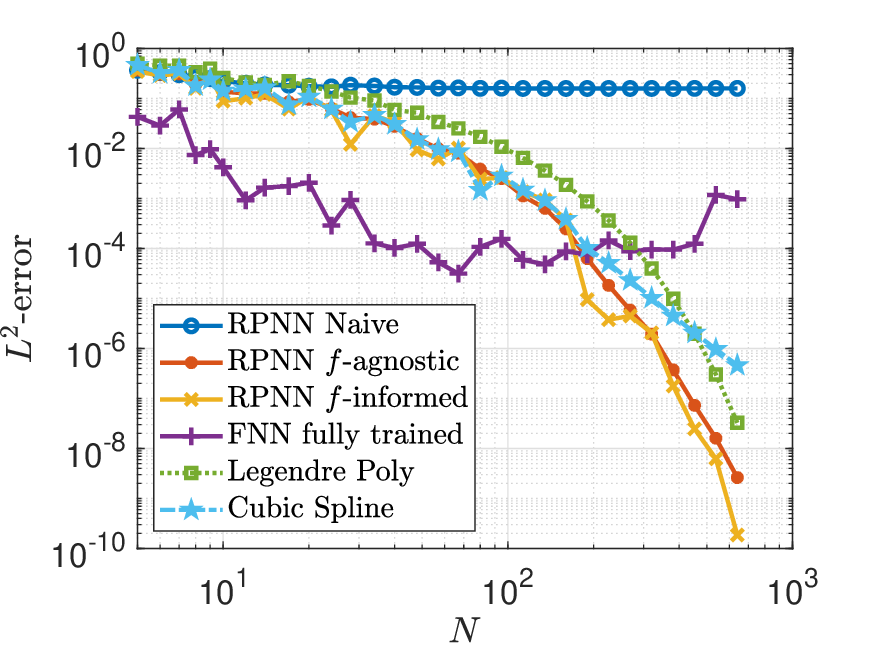}
    }
    \caption{First benchmark example, function $f_1$ in Eq. \eqref{eq:example1}, with $k=10$ and $k=100$, presenting a high steep gradient. (a) reference functions; (b)-(c) Convergence diagrams of $L^2$-norms, for $k=10$ and $k=100$, respectively. We compare Legendre Polynomial, RPNNs of best $L^2$ approximation, standard FNN trained with Levenberg-Marquardt algorithm and Cubic Spline. For the RPNNs we compare 3 different selection (naive, function-agnostic and function-informed) of the internal parameters as explained in Section \ref{sec:selection}. For the RPNNs we report the mean accuracy out of 100 different Monte-Carlo selections of the internal parameters. For the FNN we report the best network out of 10 runs with different initialization of the weights. \label{fig:example1}}
\end{figure}

For all examined function approximators, i.e. either polynomials, splines or networks, say $q_N$ of a function $f$, we report the $L^2$-error given by:
\begin{equation}
    \|f-q_N\|_2=\biggl(\int_{a}^b (f(x)-q_N(x))^2\biggr)^{\frac{1}{2}}
\end{equation}
where the integral is approximated with a composite trapezoidal quadrature rule over a dense grid of $10,000$ equally spaced points (test set).

\subsection{Case study 1: functions with steep gradient}
For our first illustration, we consider the approximation of the function\cite{calabro2021extreme}:
\begin{equation}
    f_1(x):= \arctan\biggl(k\big(x-\frac{4}{9}\big)\biggr) \qquad x \in [0,1],
    \label{eq:example1}
\end{equation}
for $k=10$ and $k=100$. The function for $k=100$ presents an internal high steep gradient resembling a discontinuity.
The reference functions and the results are depicted in Figure \ref{fig:example1}. We report the $L^2$-error computed on a dense grid of $10,000$ equally spaced points in the interval $[0,1]$.
As shown in Figure \ref{fig:example1}, Both function-agnostic and function-informed RPNNs demonstrate remarkable accuracy, resulting comparable with exponential convergence of Legendre polynomials, while significantly outperforming cubic spline interpolation. However, despite the theoretical findings, naive random selection in RPNNs plateaus around a mere 2 digits of accuracy. 
While fully trained FNNs exhibit decent performance with a limited number of neurons in the hidden layers, their accuracy quickly plateaus around 4 digits of accuracy and struggles to reach high precision.

\begin{figure}[ht!]
    \centering
    \subfigure[references]{\includegraphics[width=0.31\textwidth]{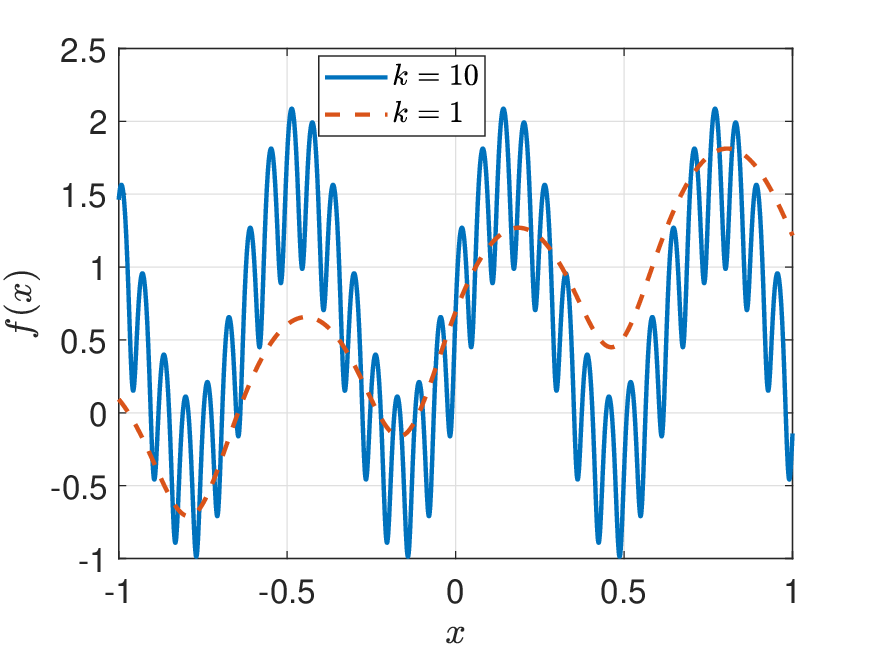}}
    \subfigure[results $k=1$]{
    \includegraphics[width=0.31\textwidth]{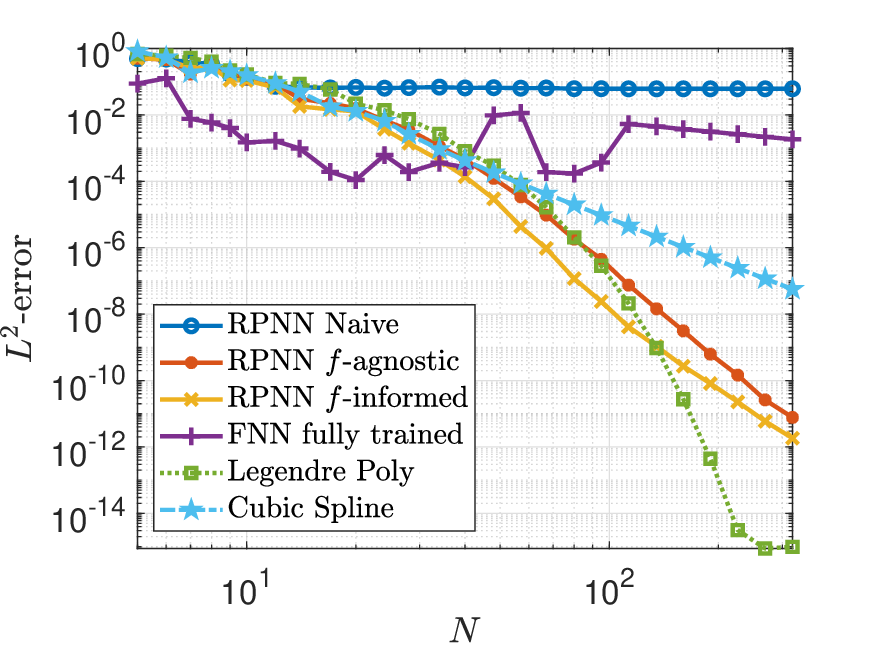}
    }
    \subfigure[results $k=10$]{
    \includegraphics[width=0.31\textwidth]{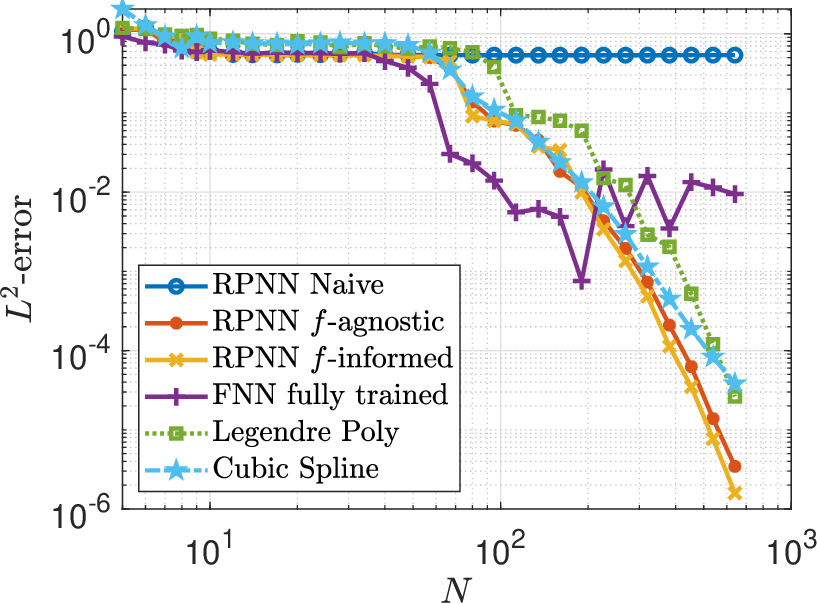}
    }
    \caption{Second benchmark example, function $f_2$ in Eq. \eqref{eq:example2}, with $k=1$ and $k=10$, presenting high-oscillations. (a) reference functions; (b)-(c) Convergence diagrams in $L^2$-norm, for $k=1$ and $k=100$, respectively. We compare Legendre Polynomials, RPNNs of best $L^2$ approximation, standard FNN and Cubic Spline. For the RPNNs we compare 3 different selection (naive, function-agnostic and function-informed) of the internal parameters as explained in Section \ref{sec:selection}. For the RPNNs we report the mean accuracy out of 100 different Monte-Carlo selections of the internal parameters. For the FNN we report the best network out of 10 runs with different initialization of the weights. \label{fig:example2}}
\end{figure}

\subsection{Case study 2: functions with high-oscillations}
Here, we consider the approximation of the function\cite{adcock2021gap}:
\begin{equation}
    f_2(x):=\log(\sin(10kx) + 2) + \sin(kx), \qquad x \in [-1,1],
    \label{eq:example2}
\end{equation}
for $k=1$ and $k=10$.
The reference functions and the results are depicted in Figure \ref{fig:example2}. As can be noted, for $k=10$ the reference function presents high frequency oscillations. We report the $L^2$-error computed on a dense grid of $10,000$ equally spaced points in the interval $[-1,1]$.
\begin{figure}[ht!]
    \centering
    \subfigure[reference at $t=1/\pi$]{\includegraphics[width=0.31\textwidth]{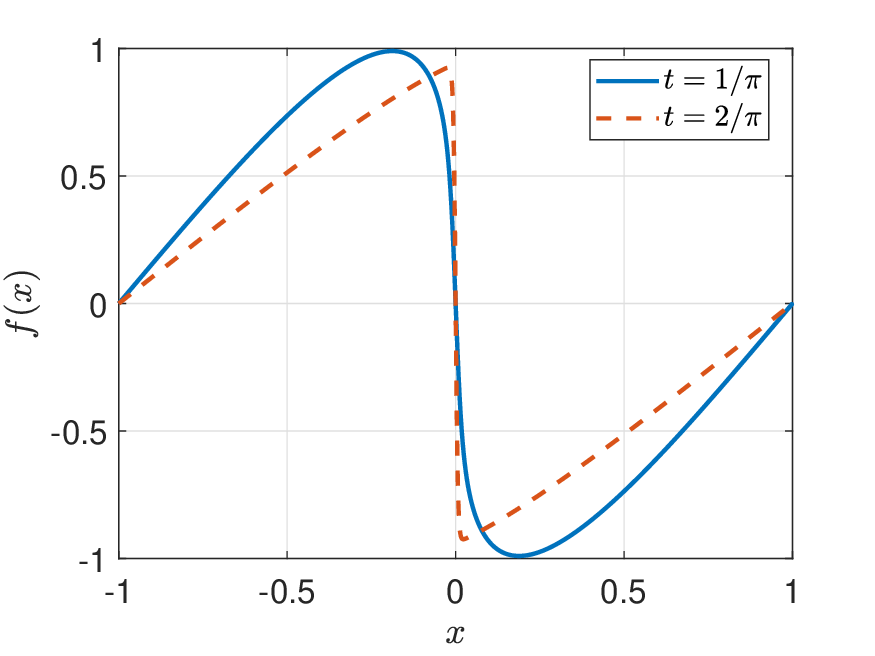}}
    \subfigure[results at $t=1/\pi$]{
    \includegraphics[width=0.31\textwidth]{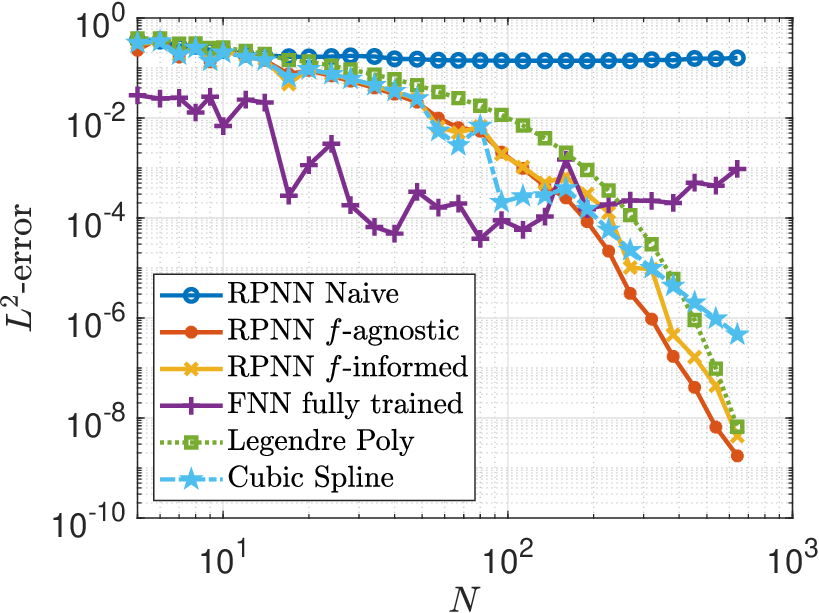}
    }
    \subfigure[results at $t=2/\pi$]{
    \includegraphics[width=0.31\textwidth]{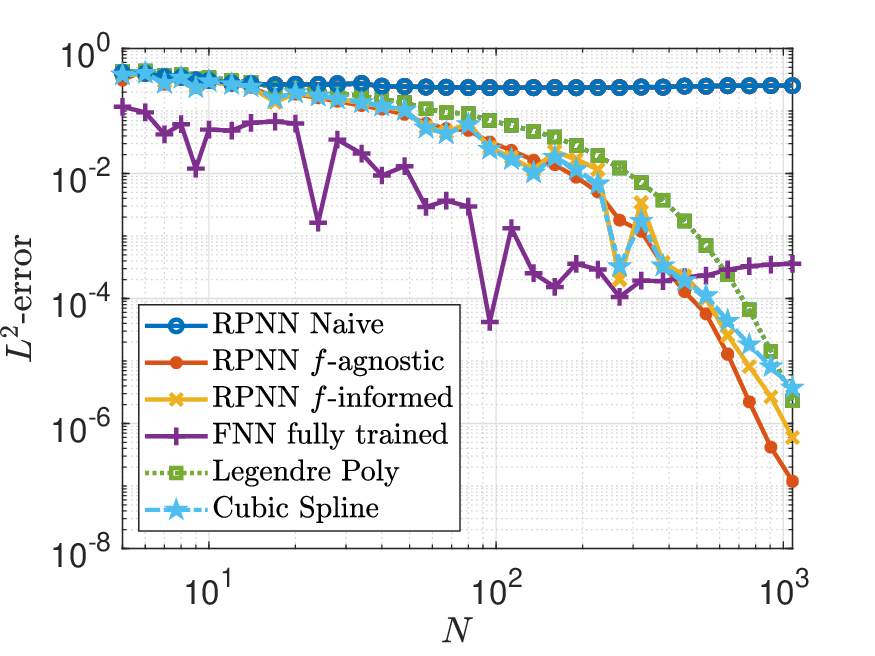}
    }
    \caption{Fourth benchmark example, approximating the analytical solution $f_3$ in Eq. \eqref{eq:burgers_sol} of the Burgers' PDE in Eq. \eqref{eq:burgersPDE} with viscosity $\nu=\frac{0.01}{\pi}$. (a) reference functions at time $t=1/\pi$ and $t=2/\pi$; (b)-(c) Convergence diagrams of $L^2$-norm, comparing the Legendre Polynomial, RPNNs of best $L^2$ approximation, standard FNN trained with Levenberg-Marquardt algorithm and Cubic Spline. For the RPNNs we compare 3 different selection (naive, function-agnostic and function-informed) of the internal parameters as explained in Section \ref{sec:selection}. For the RPNNs we report the mean accuracy out of 100 different Monte-Carlo selections of the internal parameters. For the FNN we report the best network out of 10 runs with different initialization of the weights. \label{fig:example4}}
\end{figure}
Figure \ref{fig:example2}, highlights the high accuracy of both function-agnostic and function-informed RPNNs, matching the exponential convergence of the Legendre polynomials up to 8 digits of accuracy. However, beyond 8 digits of accuracy, RPNNs seems to transition towards a less efficient high-order algebraic convergence (greater than cubic splines), suggesting potential ill-conditioning issues during training.
In this case, for high-oscillations, naive random selection RPNNs fails and plateaus around only 1 digits of accuracy. Meanwhile, the performance of fully trained FNNs exhibit decent performance in adapting its basis to the high-oscillations, yet struggling to reach more than 4 digits of accuracy.

\subsection{Case study 3: approximating infinite series function (Burgers' PDE solution)}
Here, we consider the approximation of the solution of the analytical solution of the Burgers' PDE, that exhibit shock waves. The PDE is given by:
\begin{equation}
    \frac{\partial u}{\partial t}=\nu \frac{\partial^2 u}{\partial x^2}-u \cdot \frac{\partial u}{\partial x}, \qquad x \in [-1,1], \qquad \nu=\frac{0.01}{\pi}
    \label{eq:burgersPDE}
\end{equation}
with homogeneous Dirichlet boundary conditions.  The initial condition is:
\begin{equation}
    u(0,x)=-sin(\pi x).
\end{equation}
The analytical solution, that was found by Cole \cite{cole1951quasi}, can be expressed as infinite series that can be approximated with numerical Hermite integration \cite{benton1972table}:
\begin{equation}
    f_3(t,x):=u(t,x)=4\pi \nu \frac{\sum_{n=1}^{\infty} n a_n \exp(-n^2 \pi^2 t\nu)\sin(n \pi x)}{a_0+2\sum_{n=1}^{\infty} n a_n \exp(-n^2 \pi^2 t\nu)\cos(n \pi x)},
    \label{eq:burgers_sol}
\end{equation}
where $a_n = (-1)I_n(\frac{1}{2\pi\nu})$ and $I_n(z)$ denotes the modified Bessel functions of first kind.
The reference function $f_3$ at time $t=1/\pi$ and $t=2/\pi$, and the results are depicted in Figure \ref{fig:example4}. We report the $L^2$-error computed on a dense grid of $10,000$ equally spaced points in the interval $[-1,1]$.
\begin{figure}[ht!]
    \centering
    \subfigure[reference]{\includegraphics[width=0.31\textwidth]{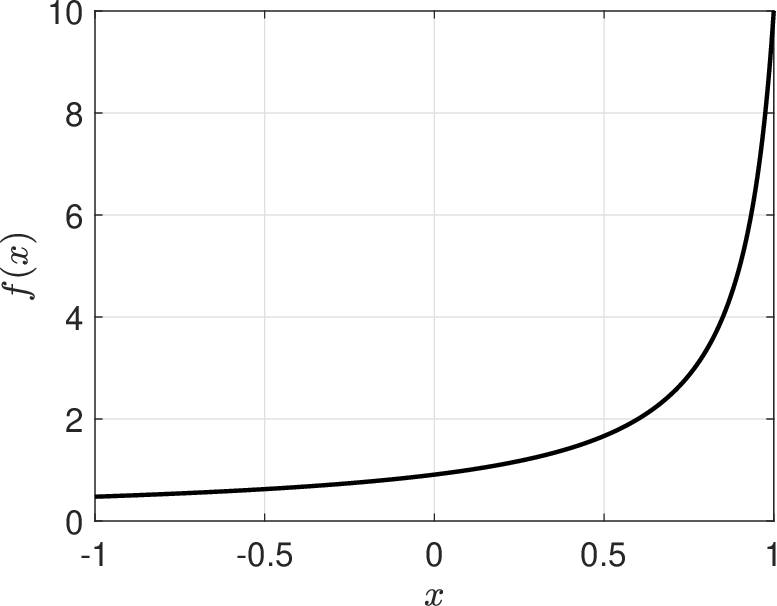}}
    \subfigure[Best $L^2$]{
    \includegraphics[width=0.31\textwidth]{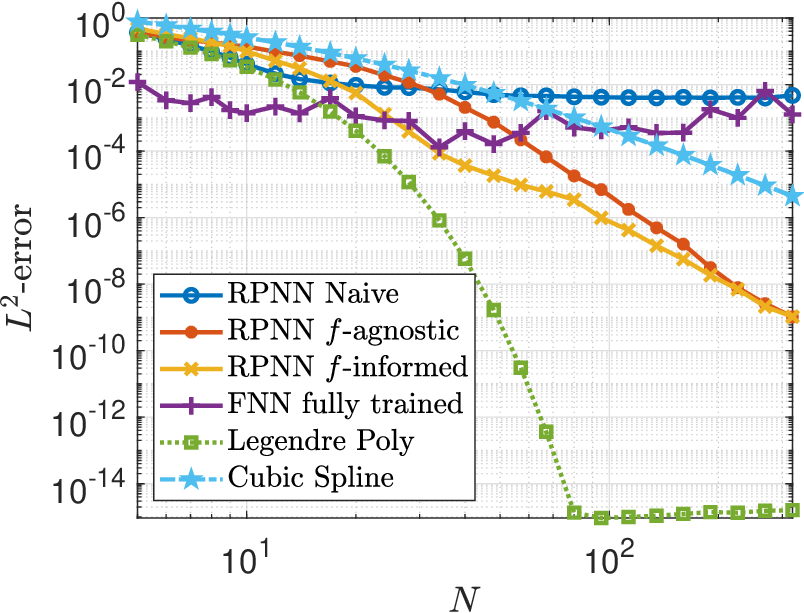}
    } \,\\
    \subfigure[reference]{\includegraphics[width=0.31\textwidth]{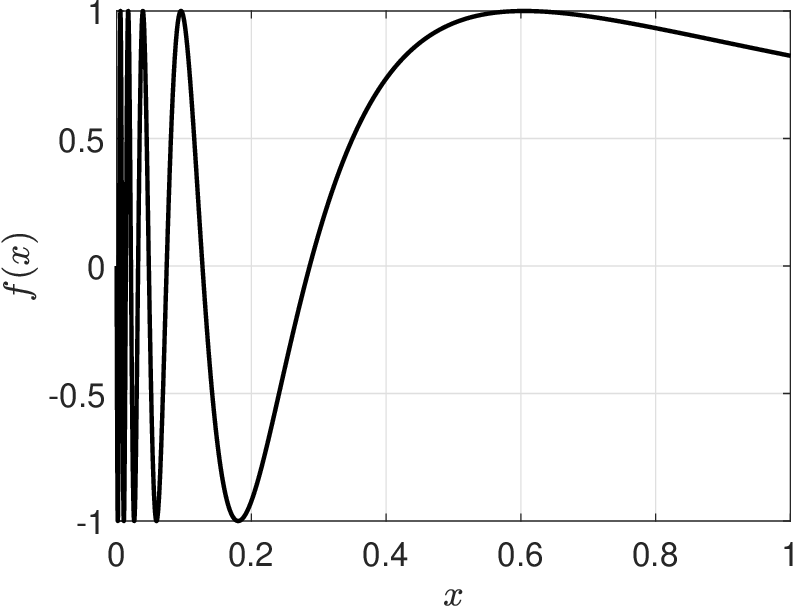}
    }
    \subfigure[Best $L^2$]{\includegraphics[width=0.31\textwidth]{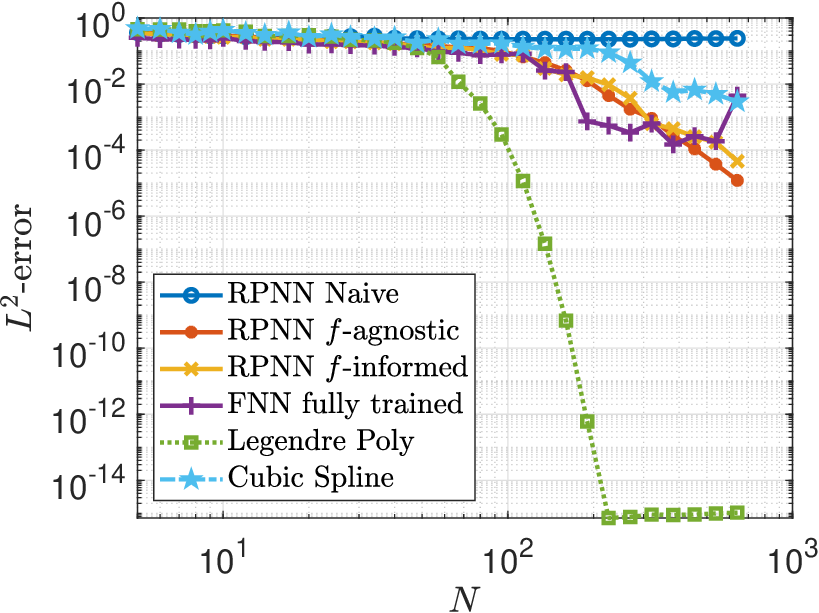}}
    \caption{Benchmark examples, function $f_4$ in Eq. \eqref{eq:example5a} and $f_5$ in Eq. \eqref{eq:example5b} approaching discontinuity. (a)-(c) reference functions; (b)-(d) Convergence diagrams $L^2$-norm comparing Legendre Polynomial, RPNNs of best $L^2$ approximation, standard FNN and Cubic Spline. For the RPNNs we compare 3 different selection (naive, function-agnostic and function-informed) of the internal parameters as explained in Section \ref{sec:selection}. For the RPNNs we report the mean accuracy out of 100 different Monte-Carlo selections of the internal parameters. For the FNN we report the best network out of 10 runs with different initialization of the weights. \label{fig:example5}}
\end{figure}
In Figure \ref{fig:example4}, it is evident that non-naive RPNNs exhibit exponential convergence as Legendre polynomials, even surpassing their performance. While Cubic splines initially demonstrate convergence comparable to RPNNs, they subsequently exhibit a slight decrease in accuracy. Conversely, fully trained FNNs achieve relatively good performances, albeit limited to around 4 significant digits of accuracy. We remark that, in addressing the Burgers' PDE, the analytical solution proves challenging to be evaluated directly, necessitating an approximation typically limited to 8 significant digits \cite{benton1972table}. Consequently, higher precision was not pursued.

\subsection{Case study 4: Numerical challenges in approximating functions near singularities}
We consider the approximation of two functions in the vicinity of a discontinuity. The first diverges approaching the discontinuity at $x=-1-\epsilon$:
\begin{equation}
    f_4(x):=\frac{1}{1+\epsilon-x}, \qquad x \in [-1,1]
    \label{eq:example5a}
\end{equation}
and the second infinitely increases its oscillations approaching the discontinuity at $x=-\epsilon$ \cite{calabro2021extreme}:
\begin{equation}
    f_5(x):=\sin\biggl(\frac{1}{x+\epsilon}\biggr), \qquad x \in [0,1], \qquad \epsilon=\frac{1}{10\pi},
    \label{eq:example5b}
\end{equation}
where for the selected the parameter $\epsilon=\frac{1}{10\pi}$, the function $f_5$ has 10 zeros and 5 oscillations in the domain $[0,1]$.
The reference functions $f_4$ and $f_5$ and the corresponding numerical results are depicted in Figure \ref{fig:example5}. For both $f_4$ and $f_5$ examples, we report the $L^2$-error computed on a dense grid of $10,000$ equally spaced points in the intervals $[-1,1$ and $[0,1]$, respectively.
The two functions are infinitely differentiable in the considered domain, thus, as can be seen in Figure \ref{fig:example5}, the Legendre polynomials converge exponentially. However, the vicinity of the singularities reduces the convergence rates of the non-naive RPNNs. Nevertheless, RPNNs ultimately reach 8 digits of accuracy for $f_4$ and 4 digits of accuracy for $f_5$, outperforming the cubic spline interpolations. While FNNs encounter greater difficulty with the diverging function $f_4$, they accommodate the high oscillations of $f_5$.

\subsection{Computational cost}
Here we summarize the computational cost of the analyzed algorithms. Lagrange barycentric interpolation for finding the coefficients of Legendre polynomials requires $O(N)$ operations, where $N$ is the order of the polynomial (see \cite{glaser2007fast}).
For the computation of the cubic spline, one has to solve a banded tridiagonal linear system, with a $O(N)$ computational cost.

In contrast, for FNNs there are no direct methods for finding the weights and biases. Also the number of iterations and the size of the network significantly impact the training time. Additionally, restarting with different initial weights (due to potential training failures) increases variability. On the other side, RPNNs offer a faster alternative to iterative-based gradient-descent algorithms. Computing the Complete Orthogonal Decomposition (COD) or the Singular Value Decomposition (SVD) of the matrix $R$ in Eq.\eqref{eq:RPNN_solve}, for an RPNN with $N$ neurons and $n$ data points, requires $O(n^2N + N^2n)$ operations, resulting at least in $O(N^3)$ complexity.

Overall, the most suitable method depends on the specific problem and priorities. Without doubt Legendre polynomials and cubic splines offer lower computational cost and are more suitable when computational time is critical. However, here our main findings demonstrate that RPNNs can achieve comparable accuracy to Legendre polynomial while being significantly more efficient than the iterative training process required for fully trained ANNs.

\section{Discussion}
Our findings in Section \ref{sec:main} demonstrate that there exists an RPNN with fixed weights and biases that can achieve an exponential convergence rate in norm $L^p$ by theoretically mimicking the best polynomial approximation. This theoretically implies at least exponential convergence for the RPNN of best $L^p$ approximation.
However, it is important to note that while the presented constructive procedure in Theorem \ref{thm:mainthm} provides a theoretical foundation for exponentially convergent approximation, it may not be considered a fast algorithm in practice, as it involves first the construction of the best $L^p$ approximation polynomial. It also needs a representation of the polynomial in the canonical base. Besides, it is necessary the inversion of the matrix $M$ in Eq. \eqref{eq:A_matrix} which may be ill-conditioned. On the other side, it is possible to seek directly for the RPNN of best $L^2$ approximation by solving a linear least squares  problem, in Eq. \eqref{eq:RPNN_solve}, by employing classical methods, such as Tikhonov regularization or, to improve robustness to the ill-conditioned training, Singular Value Decomposition (SVD) in Eq. \eqref{eq:pseudo_inverse} and Complete Orthogonal Decomposition (COD) in Eq. \eqref{eq:COD}.
For illustration and test purposes, we considered five benchmark problems for function approximation, including functions with challenging features like steep gradients, high oscillations, near-discontinuities, and approximations of infinite series function. These benchmarks showcase the versatility and applicability of the RPNNs approach. However, the results reveal the necessity of design (a) parsimonious choices of the basis functions as well as (b) a robust treatment of the ill-conditioned training process.
Indeed, while our theoretical results guarantee exponential convergence, numerical experiments revealed that a naive random selection of RPNN's internal parameters falls short of this potential. To these extent, for (a) we have discussed three different strategies for the pre-selection of the internal weights in Section \ref{sec:selection} and for (b) we compared the inversion of the matrix $R$ based on SVD or COD in Appendix \ref{sec:SVDvsCOD}. We proved that the combination of function-informed RPNNs, as in Section \ref{sec:random_informed_selection}, and COD-based solution of \eqref{eq:RPNN_solve}, results in an effective numerical exponential converge comparable with Legendre polynomials, and ultimately, in most cases, RPNNs can achieve a numerical precision up to 14 digits of accuracy. 
However, a limitation emerged near diverging discontinuities, e.g. when dealing with functions exhibiting ``exploding" behavior or rapidly increasing oscillations, as in benchmark function examples $f_4$ and $f_5$, respectively. Future research should investigate strategies to address this specific challenges and further expand the practical applicability of ANNs and RPNNs.
Besides, extending the function-informed selection to high-dimensional problems remains a challenge.\par

\par

Finally, while RPNNs might exhibit comparable accuracy in approximating functions, current evidence suggests that neither RPNNs nor FNNs offer significant advantages over Legendre polynomials in terms of accuracy or computational efficiency for the considered low-dimensional problems. Nonetheless, our research revealed that RPNNs offers a robust, accurate and efficient alternative to iteratively trained FNNs. While beyond the scope of this study, future work could explore the use of RPNNs in cases where neural networks potentially offer advantages over polynomial-based approaches. 
In particular, it is worth noting that a FNN with the same architecture as an RPNN theoretically have the potential for superior accuracy.
Yet, the current state-of-art iterative algorithms cannot reach a precision beyond 4 digits of accuracy.
Hence, it is not trivial to realize that the optimized basis of fully trained FNNs could be less effective than (non-naive) randomly uniformly initialized ones.
To explain this discrepancy, we can imagine that the fully trained FNNs gradually adapt and converge towards a specific set of basis functions along with their corresponding read-out external weights. During this process, the presence of numerous local minima poses challenges, with gradient descent algorithms often becoming stuck to less effective basis functions.
Therefore, further research on algorithms for the optimization of the basis functions is crucial to bridge this gap and improve their performance. We believe that by delving deeper into the function-informed selection, we can gain a more comprehensive understanding of the factors influencing the accuracy limitations of FNNs and identify potential strategies to overcome them, ultimately contributing to the advancement of both FNNs and RPNNs.

\section*{Declarations}
\subsection*{Acknowledgements}
The author expresses gratitude to the Professor Constantinos Siettos for his invaluable assistance and engaging discussions, which have significantly enriched the content and quality of this paper.

\subsection*{Conflict of Interests}
The author hereby declares that there is no conflict of interest associated with the research presented in this paper.

\subsection*{Data and Code Availability}
The MATLAB code used in this study, including scripts and algorithms specific to the simulations described in the case studies, will be made available upon reasonable request.

\appendix

\section{Proof of interpolation property of RPNNs}
\label{sec:proof_interpolation}
We prove here the Theorem \ref{thm:huang} about exact interpolation of RPNNs.
\begin{theorem}[Exact interpolation of RPNNs]
\begin{proof}
    We will prove that, the matrix $R$ in Eq. \eqref{eq:RPNN_solve} is invertible with probability 1.
    Consider the $j$-th column vector $\mathbf{v}(\beta_j) = \left[ \psi_j(\bm{x}_0), \ldots, \psi_j(\bm{x}_N) \right]^T = \left[ \psi(\bm{\alpha}_j \bm{x}_1 + \beta_j), \ldots, \psi(\bm{\alpha}_j \bm{x}_N + \beta_j) \right]^T$ of matrix $R$ in Euclidean space $\mathbb{R}^N$, where $\beta_j \in (\beta_{min},\beta_{max}) \subset \R$ is the corresponding internal bias.
    We want to establish that the vector $\mathbf{v}$ does not belong to any subspace with dimension less than $N+1$.
    Since $\bm{x}_k\neq\bm{x}_{k'}$ are distinct, and since the weights A are drawn from a continuous probability distribution, we can assume with probability 1 that also $\bm{\alpha}_j \bm{x}_k\neq \bm{\alpha}_j \bm{x}_{k'}$, for all $k\neq k'$. Also we get with probability 1 that $(\bm{\alpha}_j,\beta_j)\neq \pm(\bm{\alpha}_{j'},\beta_{j'})$, thus satisfying the condition of Ito \cite{ito1996nonlinearity} to obtain a system of independent (plane wave) basis functions. If by contradiction, we assume that $\mathbf{v}$ belongs to a subspace of dimension $N$ (less than $N+1$). Then, there exists a vector $\mathbf{u}=(u_0,u_1,\dots,u_N)\neq \bm{0}$ which is orthogonal to this subspace and such that:
    \begin{equation}
        \mathbf{u} \cdot (\mathbf{v}(\beta_j) - \mathbf{v}(\beta_{min})) = u_0 \psi(\beta_j + z_0) + u_1 \psi(\beta_j + z_1) + \ldots + u_N \psi(\beta_j + z_N) - \mathbf{u}\cdot \mathbf{v}(\beta_{min}) = 0,
    \end{equation}
where $z_k = \bm{\alpha}_j \bm{x}_k$ for $k = 0, \ldots, N$, for all $\beta_j \in (\beta_{min},\beta_{max})$. Assuming, without loss of generality, $u_N \neq 0$, the above equation can be further expressed as:
\begin{equation}
    \psi(\beta_j + z_N) = \sum_{l=0}^{N-1} \gamma_l \psi(\beta_j + z_l) + \frac{\mathbf{u}\cdot \mathbf{v}(\beta_{min})}{u_N},
\end{equation}
where $\gamma_l = \frac{u_l}{u_N}$ for $l = 0, \ldots, N-1$. Utilizing the fact that $\psi$ is infinitely differentiable and a non-polynomial, we have:
\begin{equation}
    \psi^{(m)}(\beta_j + z_N) = \sum_{l=0}^{N-1} \gamma_l \psi^{(m)}(\beta_j + z_l), \qquad m=1, 2, \ldots, N, N+1,\ldots
    \label{eq:proof_thm_huang1}
\end{equation}
However, there are only $N$ free coefficients $\gamma_0, \gamma_1, \ldots, \gamma_{N-1}$, for the resulting (more than) $N+1$ linear equations in Eq. \eqref{eq:proof_thm_huang1}. Indeed, we can note that the matrix $W$ with entries $W_{m,l}=\psi^{(m)}(\beta_j+z_l)$ is a Wronskian matrix. Since we have a basis of linear independent functions with probability 1, the Wronskian matrix of size $N+1 \times N+1$ is invertible. But Eq. \eqref{eq:proof_thm_huang1} implies that only $N$ columns are independent.
This contradiction implies that the vector $\mathbf{v}$ does not belong to any subspace with a dimension less than $N+1$.
\end{proof}
\end{theorem}

\section{Comparison of SVD and COD for the ill-conditioned training of RPNNs}
\label{sec:SVDvsCOD}
Here we perform some additional numerical tests, for the performance of the RPNNs with function-agnostic and function-informed selections of the weights, when solving the minimum norm least square problem in \eqref{eq:minimum_norm}.
\begin{figure}[ht!]
    \centering
    \subfigure[$f_1$ with $k=10$]{\includegraphics[width=0.31\textwidth]{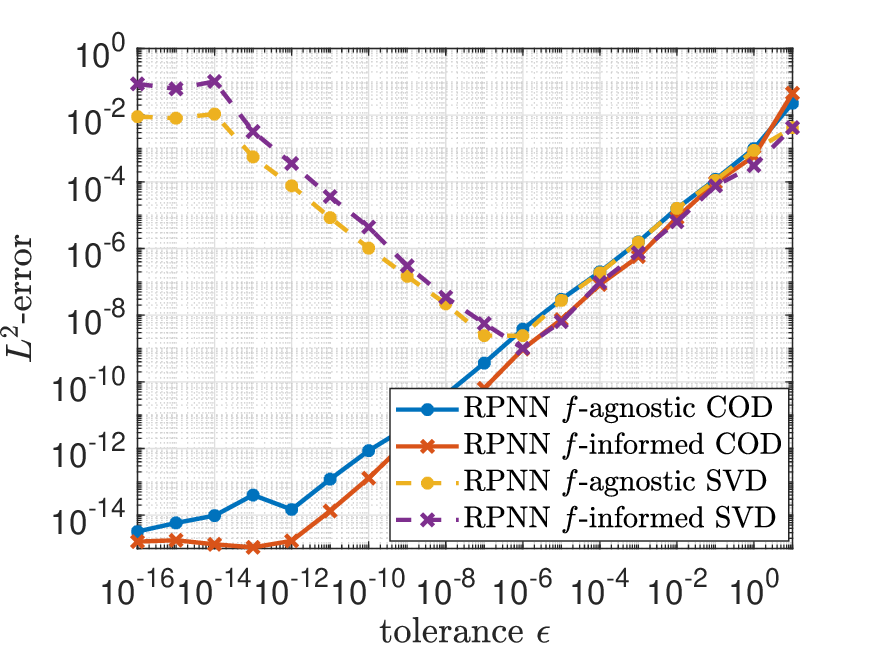}}
    \subfigure[$f_1$ with $k=100$]{\includegraphics[width=0.31\textwidth]{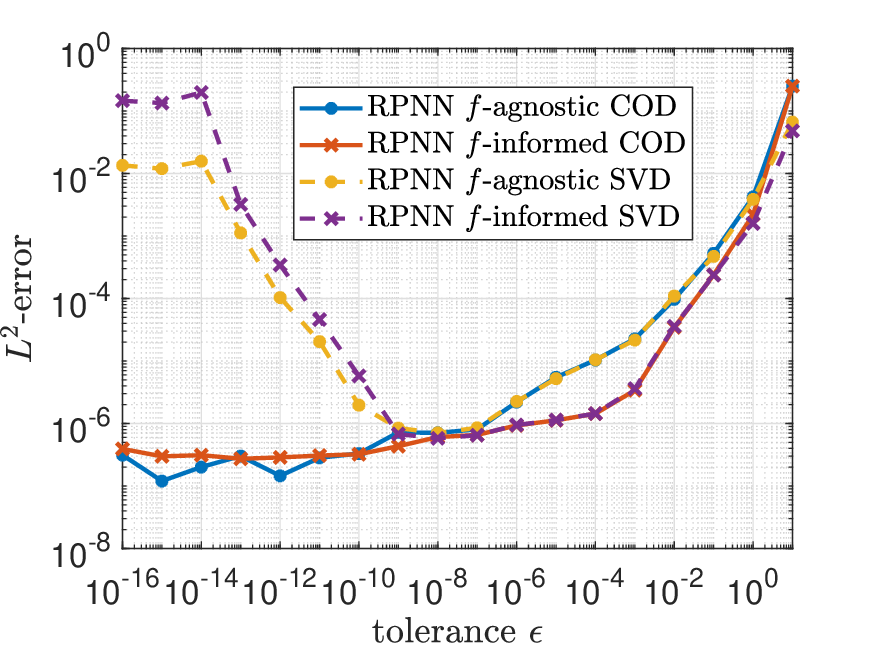}}\,\\
    \subfigure[$f_2$ with $k=1$]{\includegraphics[width=0.31\textwidth]{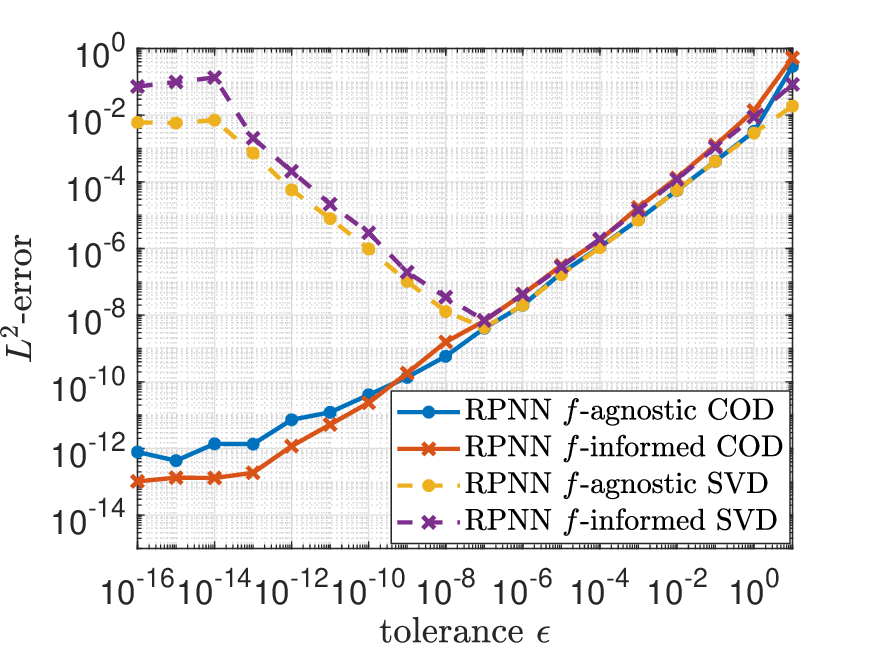}}
    \subfigure[$f_2$ with $k=10$]{\includegraphics[width=0.31\textwidth]{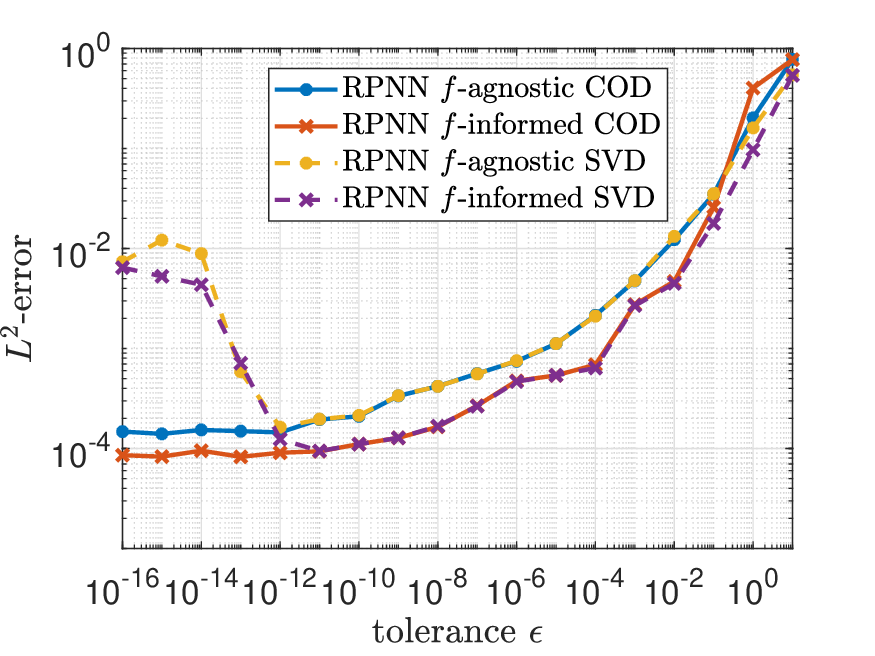}}
    \caption{Comparison of Singular Value Decomposition (SVD)-based and Complete Orthogonal Decomposition (COD)-based solution of the problem \eqref{eq:minimum_norm}. We fix the number of neurons to $N=400$.  Convergence with respect to the tolerance $\epsilon$ for (a) $f_1$ in Eq.\eqref{eq:example1} with $k=10$, (b) $f_1$ in Eq.\eqref{eq:example1} with $k=100$, (c) $f_2$ in Eq.\eqref{eq:example2} with $k=1$,  (d) $f_2$ in Eq.\eqref{eq:example2} with $k=10$, respectively. \label{fig:svd_cod_comparison}}
    
\end{figure}
In particular, here we are interested in showing the role of the tolerance $\epsilon$ and compare the SVD-based and COD-based pseudo-inversion of the matrix $R$ in Eq. \eqref{eq:RPNN_solve}. In Figure \ref{fig:svd_cod_comparison}, we depict a comparison of the SVD vs the COD, for the accuracy performance of both RPNNs with function-agnostic and function-informed selections of the internal weights. In particular, we fixed the number of neurons to $N=400$ and we made vary the value of tolerance $\epsilon>0$ used the two solving methods. We trained the RPNNs for the benchmark functions $f_1$ in Eq. \eqref{eq:example1} and $f_2$ in Eq. \eqref{eq:example2}. Also in this case we perform 100 different Monte-Carlo selections of the weights and we report the mean accuracy.
As can be seen in Figure \ref{fig:svd_cod_comparison}, the COD-based solution is more stable for rank-deficient least square minimum norm training of RPNNs as in Eq. \eqref{eq:minimum_norm}, allowing to obtain, working in double precision, up to 14 digits of accuracy. While SVD-based solution seems to be stable up to a tolerance $\epsilon=$1E$-$08, obtaining up to 8 digits accuracy.

\bibliographystyle{abbrv}
\bibliography{AA_references}

\end{document}